\pdfoutput=1

\documentclass[11pt]{article}

\usepackage{acl}

\usepackage{times}
\usepackage{latexsym}

\usepackage[T1]{fontenc}

\usepackage[utf8]{inputenc}

\usepackage{microtype}

\usepackage{inconsolata}

\usepackage{graphicx}
\usepackage{xspace}
\usepackage{booktabs}
\usepackage{amsfonts}
\usepackage{amsmath}
\usepackage{kotex}
\usepackage{xcolor}
\usepackage{subcaption}
\usepackage{multirow}
\usepackage{tikz}
\usepackage{enumitem}

\newcommand{\sys}{\mbox{\textsc{PEMA}}\xspace}

\newcommand*\BC[1]{%
\begin{tikzpicture}[baseline=(C.base)]
\node[draw,circle,fill=black,inner sep=0.2pt](C) {\textcolor{white}{#1}};
\end{tikzpicture}}

\title{PEMA: An Offsite-Tunable Plug-in External Memory Adaptation for Language Models}

\author{HyunJin Kim \\ Sungkyunkwan University \\ Suwon, South Korea \\ \texttt{khyunjin1993@skku.edu} 
\And
Young Jin Kim$^{^*}$ \\ Microsoft \\ Redmond, USA \\  
\texttt{youki@microsoft.com} \\
\And
JinYeong Bak$^{^*}$ \\ Sungkyunkwan University \\ Suwon, South Korea \\
\texttt{jy.bak@skku.edu}}

\begin{document}
\maketitle
\begingroup\def\thefootnote{*}\footnotetext{Corresponding authors}\endgroup
\renewcommand{\thefootnote}{\arabic{footnote}}
\begin{abstract}
Pre-trained language models (PLMs) show impressive performance in various downstream NLP tasks.
However, pre-training large language models demands substantial memory and training compute.
Furthermore, due to the substantial resources required, many PLM weights are confidential. 
Consequently, users are compelled to share their data with model owners for fine-tuning specific tasks.
To overcome the limitations, we introduce Plug-in External Memory Adaptation (\sys), a Parameter-Efficient Fine-Tuning (PEFT) method, enabling PLM fine-tuning without requiring access to all the weights. 
\sys integrates with context representations from test data during inference to perform downstream tasks.
It uses external memory to store PLM-generated context representations mapped with target tokens. 
Our method utilizes weight matrices of LoRA-like bottlenecked adapter in the PLM's final layer to enhance efficiency.
Our approach also includes Gradual Unrolling, a novel interpolation strategy to improve generation quality.
We validate \sys's effectiveness through experiments on syntactic and real datasets for machine translation and style transfer. 
Our findings show that \sys outperforms other PEFT approaches in memory and latency efficiency for training, and also excels in maintaining sentence meaning and generating appropriate language and styles.
\end{abstract}
\section{Introduction}
\label{s:introduction}
\begin{figure}[t!]
    \centering
    \begin{subfigure}[b]{0.47\textwidth}
        \centering
        \includegraphics[width=1.0\linewidth]{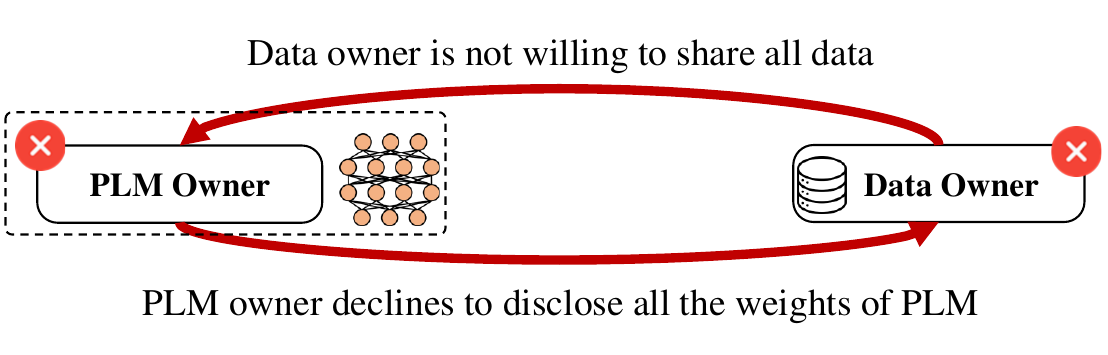}
        \caption{Problems for fine-tuning proprietary PLMs}
        \label{fig:motivation_1}
    \end{subfigure}
    \begin{subfigure}[b]{0.47\textwidth}
        \centering
        \includegraphics[width=1.0\linewidth]{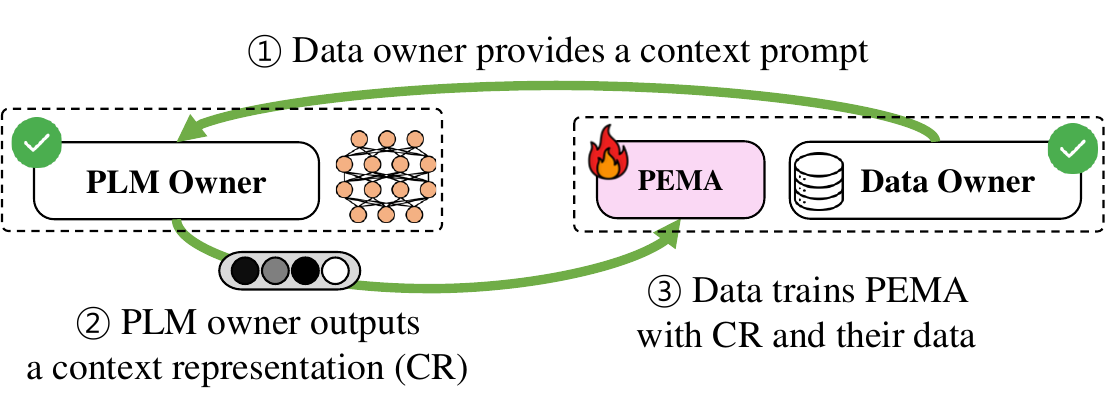}
        \caption{\sys training phase}
        \label{fig:motivation_2}
    \end{subfigure}
    \begin{subfigure}[b]{0.47\textwidth}
        \centering
        \includegraphics[width=1.0\linewidth]{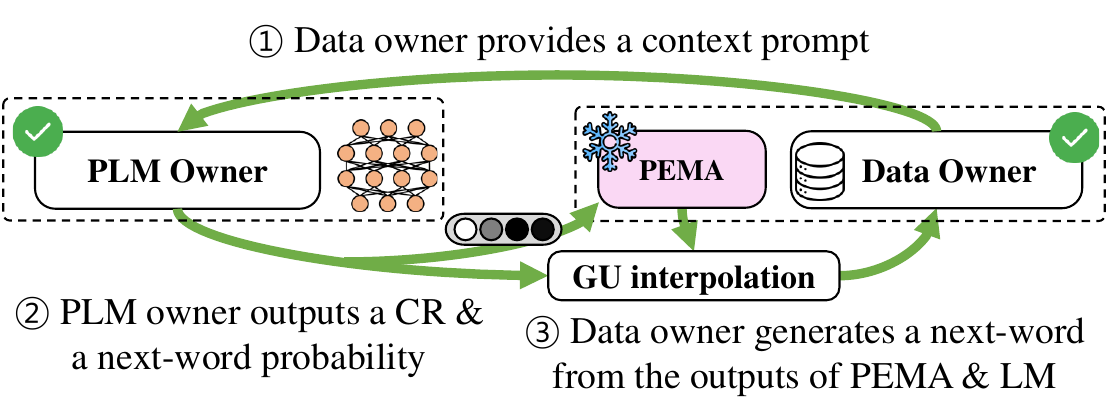}
            \caption{\sys inference phase}
        \label{fig:motivation_3}
    \end{subfigure}
    \caption{
    A motivation for \sys. 
    (a) The data owners who want to fine-tune PLMs encounter a problem when the PLM owner refuses to share all the weights of the PLM.
    (b) In the \sys training phase, the data owner takes a CR from the PLM owner by providing a context prompt. They subsequently train their \sys model with their dataset.
    (c) At inference, the data owner takes a CR for test data from the PLM owner. Using Gradual Unrolling (GU), they generate the next-token by interpolating between \sys and PLM next-token probabilities.
    }
    \label{fig:motivation}
\end{figure}
Pre-trained language models (PLMs) are widely used in downstream NLP tasks~\cite{bert}. Recent developments in large language models have shown remarkable performance in zero-shot and few-shot learning scenarios~\cite{brown2020language, gpt-mt-2023, openai2023gpt4, anil2023palm, chowdhery2022palm}. 
However, fine-tuning is still required to optimize the performance of the NLP tasks such as machine translation~\cite{ustun-2022-parameter, 8733017, ding2022delta}.
The most straightforward approach to fine-tuning is full fine-tuning~\cite{raffel2020exploring, qiu2020pre}, which involves fine-tuning all parameters in a PLM.
Yet, this approach requires substantial resources regarding memory and training compute~\cite{iyer2022opt, zhang2022opt, touvron2023llama}.
To overcome this limitation, researchers have proposed Parameter-Efficient Fine-Tuning (PEFT) methods to fine-tune a full model efficiently.
Adapter tuning~\cite{pfeiffer-etal-2021-adapterfusion, he2021towards, pmlr-v97-houlsby19a} utilizes small, additional parameters known as adapters inserted between layers within a PLM.
On the other hand, LoRA~\cite{hu2022lora} uses trainable low-rank matrices that incrementally update the pre-trained weights.
These fine-tuning methods require access to all the weights of PLMs.

However, proprietary PLMs such as ChatGPT~\cite{chatgpt}, Bard~\cite{bard}, and Claude~\cite{claude} are confidential. Hence, the owners of these PLMs do not reveal all the model weights.
Consequently, data owners possessing their datasets and wishing to fine-tune proprietary PLMs for specific downstream tasks must provide their datasets to the PLM owners for fine-tuning~\cite{openaift}.
However, this process can be challenging due to the confidential nature of the datasets, which may involve privacy concerns~\cite{guinney2018alternative}. Figure~\ref{fig:motivation_1} shows problems for fine-tuning proprietary PLMs.
To overcome this situation, \cite{xiao2023offsite} proposes the offsite-tuning approach that uses one-third of the middle layers of a PLM, referred to as the emulator.
Nevertheless, this approach still needs a large parameter size, and compressing the full model into an emulator requires a computationally intensive distillation process. 

To address the challenges mentioned above, we introduce a novel PEFT method named Plug-in External Memory Adaptation (\sys) designed for efficient fine-tuning of proprietary PLMs in machine translation tasks.
{\sys utilizes weight matrices of LoRA-like bottlenecked adapter designed for learning downstream tasks}
{with accessible features provided by OpenAI API~\cite{chatgpt} and minimal part of PLM's weight (language model head)}.


In the training phase, the data owner begins the process by providing a prompt with initial input to the PLM owner, which includes an instruction and a source sentence from a parallel corpus.
The PLM owner receives this initial input to generate a context representation (i.e., a hidden representation from PLM) and predict the next-token.
Then, it iteratively processes subsequent inputs containing the predicted next-tokens. 
This approach avoids the need for the full dataset from the data owner. 
Throughout this process, the data owner builds an external memory comprised of context representations and corresponding desired target tokens. They train \sys by reconstructing the stored context representations and predicting target tokens based on these representations.
Figure~\ref{fig:motivation_2} shows the training phase process of \sys.


During the inference phase, the data owner uses a prompt to request a context representation for test data from the PLM owner. The PLM owner then outputs a context representation and a next-token probability given the prompt. \sys also outputs a next-token probability based on a context representation.
These probabilities are interpolated to compute a final next-token probability. We propose Gradual Unrolling ($GU$), an interpolation strategy that initially emphasizes \sys's distribution, gradually shifts to the PLM's context-based predictions as the sentence progresses. Figure~\ref{fig:motivation_3} illustrates the inference phase process of \sys.

We evaluate \sys by comparing it with other PEFT methods.
\sys shows better resource efficiency, consuming less GPU memory and running faster. 
Additionally, \sys outperforms other baselines in translating English sentences into German and paraphrasing informal sentences into formal ones while preserving the original meaning.
Lastly, we conduct ablation studies to assess the effectiveness of each component of \sys. \sys is publicly available for further exploration into offsite-tunable efficient fine-tuning.\footnote{\url{https://github.com/agwaBom/PEMA}}
\section{Related Work}
\subsection{Parameter-Efficient Fine-Tuning}
Parameter-Efficient Fine-Tuning aims to fine-tune PLMs to address resource constraints in memory and training compute~\cite{iyer2022opt, zhang2022opt, touvron2023llama}.
Several approaches have been proposed to overcome this limitation.
Adapter tuning~\cite{pfeiffer-etal-2021-adapterfusion, he2021towards, pmlr-v97-houlsby19a} inserts small parameters, known as adapters, between layers within a PLM.
Prefix and Prompt tuning~\cite{li-liang-2021-prefix, liu2021p, lester-etal-2021-power} incorporate additional trainable prefix tokens to a PLM's input or hidden layers.
Low-Rank Adaptation (LoRA)~\cite{hu2022lora} uses trainable low-rank matrices, denoted as $B$ and $A$, that incrementally update PLM weights. $B$ and $A$ are reduced to a low-rank $r$.
This adaptation can be mathematically represented as transitioning from $h=W_0x$ to $h=W_0x+\Delta Wx=W_0x+BAx$, where $W_0 \in \mathbb{R}^{k\times d}$, $B \in \mathbb{R}^{k\times r}$, and $A \in \mathbb{R}^{r\times d}$.
UniPELT~\cite{mao2022unipelt} combines multiple PEFT methods, using a gating mechanism to activate the most suitable components for given data or tasks. We propose a novel adaptation method that leverages a LoRA-like bottlenecked adapter\footnote{We explicitly use the term "LoRA-like bottlenecked adapter" because our method applies the parameter of LoRA on the top rather than beside the PLM's weight.} and is offsite-tunable.

\subsection{Offsite-Tuning}
Offsite-Tuning~\cite{xiao2023offsite} is designed to fine-tune proprietary PLMs while ensuring the privacy of both PLM and data owners. The process comprises three phases: emulator compression, fine-tuning, and plug-in. During the emulator compression phase, knowledge distillation is applied to reduce the PLM to one-third of its original size. The emulator is then shared with the data owner for fine-tuning using an adapter. The adapter consists of several duplicated PLM layers positioned at the beginning and end of the emulator. Throughout the fine-tuning stage, the emulator is kept frozen, and only the adapter undergoes training. Once fine-tuning is complete, the adapter is integrated back into the PLM for inference. Despite its privacy benefit, the process of Offsite-Tuning still requires a large parameter size, and compressing the full model into an emulator requires a computationally intensive distillation process. To address this problem, we propose a novel PEFT method that leverages a LoRA-like bottlenecked adapter that is efficient and offsite-tunable.

\subsection{$k$-Nearest Neighbors Language Model}
The $k$-Nearest Neighbors Language Model ($k$NN-LM) estimates the next-token distribution by interpolating the output distributions from a pre-trained language model ($P_{LM}$), and an external memory ($P_{kNN}$)~\cite{khandelwal20generalization}.
The memory is used to perform a $k$NN search and to integrate out-of-domain data, thereby enabling a single language model to be adaptive across various domains.
Given a context represented as a sequence of tokens $c_i=(w_1, ..., w_{i-1})$, the $k$NN-LM utilizes a pre-trained language model $f(\cdot)$ to generate a context representation $f(c_i)$.
This representation is then paired with the desired target token $y_i$ to create the external memory (referred to as a datastore in ~\cite{khandelwal20generalization}) $\{(f(c_i),y_i)|(c_i,y_i)\in\mathcal{E}\}$ from the training dataset $\mathcal{E}$.
The next-token distribution from the external memory, $P_{kNN}$, is computed using a $k$-nearest neighborhood approach with the squared $L^{2}$ distance.
The final next-token distribution is then obtained by interpolating between $P_{kNN}$ and $P_{LM}$ as: $P(y_i|c_i)=\lambda P_{kNN}(y_i|c_i)+(1-\lambda) P_{LM}(y_i|c_i)$.

We adapt the concept of external memory and interpolation of different next-token distributions to \sys. 
Instead of employing a $k$NN-based approach, we employ a neural network-based model that directly learns to estimate the next-token, which is more effective in mitigating overfitting to the training data. 
Additionally, we use the Gradual Unrolling interpolation strategy to enhance the quality of interpolation. 
The $k$NN-LM method relies on $k$NN for external memory search to adapt the language model to diverse domains. 
However, it is well known that the non-parametric model $k$NN can potentially overfit, especially in cases of high-dimensional input~\cite{khandelwal2021nearest, pestov2013k}.
Therefore, it often requires a large amount of training data to achieve robust performance across unseen data.
To address this, we introduce a parametric approach within \sys to improve its performance on downstream tasks. 
This approach is better suited for limited training data scenarios because a parametric approach can implement regularization to mitigate overfitting~\cite{loshchilov2018decoupled}. 
It involves replacing the existing $k$NN with a parametric model in \sys, thus enabling effective adaptation to various domains in terms of performance.

\section{Plug-in External Memory Adaptation}
\begin{figure*}[t]
    \centering
    \includegraphics[width=0.98\textwidth]{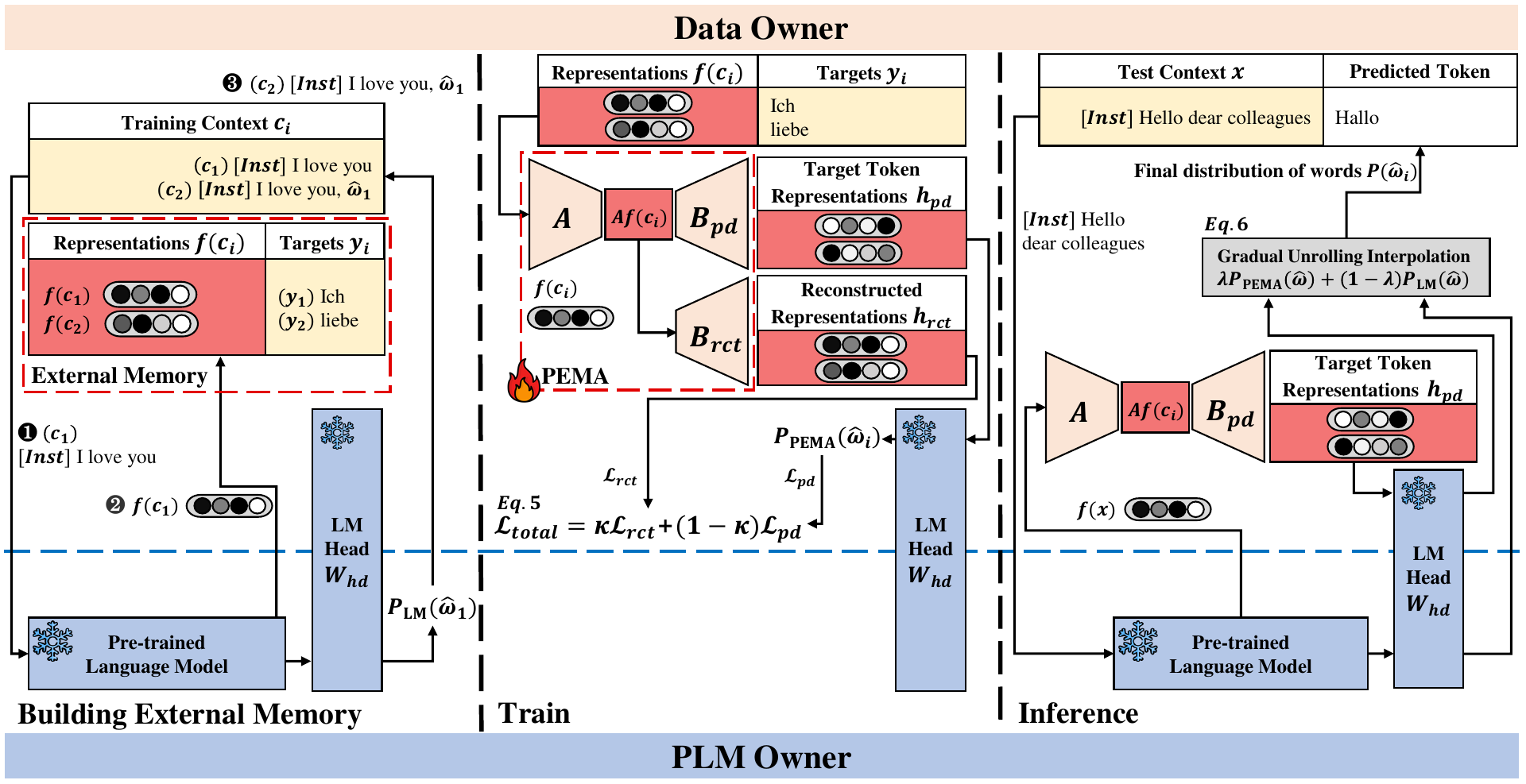}
    \caption{
    {An illustration of \sys. The areas of the PLM owner and the data owner are separated by the blue horizontal line. The data owner can train and infer using only the PLM's LM head. \sys builds an external memory from the training context with an instruction $[Inst]$ given to a PLM. 
    The PLM outputs the representation $f(c_i)$ and predicts the next-token distribution $P_{LM}(\hat{w}_i)$. The representation $f(c_i)$ is then aligned with its target $y_i$. 
    In the training phase, \sys uses external memory for two tasks: preserving the original representation via reconstruction training with $B_{rct}$ and generating a target token probability distribution using $B_{pd}$. 
    For inference, the model inputs a test data representation to generate two probability distributions: $P_{LM}(\hat{w}_i)$ and $P_{PEMA}(\hat{w}_i)$. These are then interpolated using Gradual Unrolling to obtain the final token distribution.}
    }
    \label{fig:model}
\end{figure*}
This section describes Plug-in External Memory Adaptation (\sys), which aims to fine-tune a PLM without requiring a full model during training. \sys integrates its output with that of the PLM (i.e., next-token probability) during inference to facilitate downstream NLP tasks. At training, \sys utilizes context representations of the PLM and its LoRA-like bottlenecked adapter. For inference, \sys requires context representation, the language model head (LM head) from the PLM, and the LoRA-like bottlenecked adapter.

It uses external memory to build a context representation $f(c_i)$, mapped with the desired target token $y_i$. 
Using the external memory, we train \sys in two phases.
The first phase involves reconstruction training to reconstruct $f(c_i)$ with $B_{rct}A$, resulting in the output of a reconstruction loss.
Subsequently, the joint retraining phase focuses on generating the next-token probability $P_{PEMA}$ that predicts target token $y_i$ given $Af(c_i)$ with $B_{pd}$. 
Simultaneously, it uses pre-trained $B_{rct}$ to retain the original feature $f(c_i)$.
During the inference stage, the next-token probabilities from both the pre-trained generative language model $P_{LM}$ and \sys $P_{PEMA}$ are interpolated to generate the next-token.
Figure~\ref{fig:model} shows the structure of \sys.

\subsection{Building an External Memory}
The first step of \sys is to build an external memory.
The output $f(c_i)$ represents a context representation obtained from the final layer's feed-forward network output of a pre-trained language model.

For the $i$-th token training example in external memory $(c_i, y_i) \in \mathcal{E}$, a paired representation is created by defining an input prompt $c_1$ and a corresponding target token sequence. 
Predicted token sequences are generated by sequentially extending the input prompt. 
\protect\BC{1} Initially, the input prompt $c_1$ is fed into the pre-trained language model, resulting in the predicted next-token $\hat{w}_1$ and \protect\BC{2} the corresponding context representation $f(c_1)$. 
\protect\BC{3} Including $\hat{w}_1$ in the input prompt extends it to the next context $c_2 = \{c_1, \hat{w}_1\}$, subsequently producing the next predicted token $\hat{w}_{2}$ and its context representation $f(c_2)$. This iterative process yields a sequence of context representations $(f(c_1), f(c_2), ..., f(c_{t} = \{c_1, \hat{w}_{1}, ..., \hat{w}_{t-1}\})$ for training, with each context $c_i$ corresponding to the $i$-th position in the token sequence and $t$ denoting the total number of tokens in a token sequence of one sentence training example.

We map the context representation $f(c_i) \in \mathbb{R}^{1\times d}$, where $d$ is the size of the context representation with the target token $y_i$, resulting in the pair $(f(c_i), y_i)$. The external memory $(f(C), Y)$ is formed by collecting all such context and token pairs constructed from the training set $\mathcal{E}$ as below:
\begin{equation}
    (f(C), Y) = \{(f(c_i), y_i)|(c_i, y_i)\in\mathcal{E}\}
    \label{eq:external_memory}
\end{equation}

\subsection{\sys Adaptation Model}
We use LoRA-like bottlenecked adapter~\cite{hu2022lora}, a low-rank parameterization adaptation known for its effectiveness in various adaptation tasks, into \sys for adapting to multiple text generation tasks. 

The \sys consists of three weight matrices: $A \in \mathbb{R}^{r \times d}$, $B_{rct} \in \mathbb{R}^{d \times r}$, and $B_{pd} \in \mathbb{R}^{d \times r}$ where $d$ is the size of the context representation and $r$ is a rank-size that $r < d$. 
Given $Af(c_i)$ where $f(c_i) \in \mathbb{R}^{1 \times d}$, $B_{rct}$ is used to reconstruct the context representation input $f(c_i)$, with the goal of approximating ${h_{rct}}_i \approx f(c_i)$, 
Additionally, $B_{pd}$ is used to produce a representation ${h_{pd}}_i$ that maximizes target token prediction when fed into the frozen weight of a language model {head (LM head)} $W_{hd} \in \mathbb{R}^{v \times d}$ where $v$ is the vocabulary size that outputs the predicted next-token $\hat{w}_{i}$.
\begin{equation}
\begin{gathered}
    {h_{rct}}_i=\Delta W_{rct}f(c_i) = B_{rct}Af(c_i) \\
    {h_{pd}}_i=\Delta W_{pd}f(c_i) = B_{pd}Af(c_i) \\
    P_{PEMA}(\hat{w}_{i}|c_i)=\text{softmax}(W_{hd}{h_{pd}}_i)
    \label{eq:mmm_model}
\end{gathered}
\end{equation}

\subsection{Model Training}
The training process consists of two distinct phases: initial reconstruction training {to preserve the general knowledge within the context representation of PLM} and subsequent joint retraining, encompassing both the reconstruction of context representations and the prediction of next-tokens. 

\noindent
\textbf{Initial Reconstruction Training.}
First, we train the decoder $B_{rct}$ by reconstructing the $i$-th original context representation of the $n$-th sentence training example $f(c_i)^n$.
We use a mean-square error loss between original input $f(c_i)^n$ and the output ${h^n_{rct}}_i$ as below:
\begin{equation}
    \label{eq:rct_loss}
    \mathcal{L}_{rct} = \frac{1}{|\mathcal{E}|}\sum_{n=1}^{|\mathcal{E}|}\sum_{i=1}^{t_n}(f(c_i)^n - {h^n_{rct}}_i)^2
\end{equation}
where $t_n$ is the number of tokens in a token sequence of n-th sentence training example and $|\mathcal{E}|$ is the size of the training dataset.

\noindent
\textbf{Joint Retraining}
After completing the initial reconstruction training, we proceed to the joint retraining phase, using the pre-trained $B_{rct}$ and randomly initialized $A$. 
Our first objective is to acquire a representation ${h^n_{pd}}_i$ that is optimized for predicting the target token $y^n_i$. We utilize a cross-entropy loss based on the softmax function of the output of $W_{hd}{h^n_{pd}}_i$ given the target token $y^n_i$ as below:
\begin{equation}
    \label{eq:pd_loss}
    \resizebox{.89\hsize}{!}{
    $
    \begin{split}
        \mathcal{L}_{pd} = -\frac{1}{|\mathcal{E}|}\sum_{n=1}^{|\mathcal{E}|} \sum_{i=1}^{t_n} y^n_i \log P_{PEMA}(y_i^{n}|W_{hd}{h^n_{pd}}_{i})
    \end{split}
    $
    }
\end{equation}
The second objective is to reconstruct the input context representation $x_i$ using the randomly initialized $A$ and pre-trained $B_{rct}$ with the reconstruction loss function as depicted in Equation~\ref{eq:rct_loss}. 
The reconstruction loss intends to retain the general knowledge obtained from the pre-trained language model while maximizing the target token prediction.
We introduce a parameter $\kappa$ that can be fine-tuned to adjust the emphasis on the objectives as below:
\begin{equation}
    \label{eq:total_loss}
        \mathcal{L}_{total} = \kappa \mathcal{L}_{rct} + (1-\kappa) \mathcal{L}_{pd}
\end{equation}

\subsection{Model Inference}
To generate the next-token $\hat{w}$, we exclude $B_{rct}$ and use $B_{pd}$ and $A$.
The PLM receives the input context $x$ from the test dataset, and
generates $f(x)$, which serves as input for two pathways. 
One pathway uses \sys's $A$ and $B_{pd}$ to create $h_{pd}$ for $x$. Subsequently, it is passed through $W_{hd}$ to produce a distribution of the next-token $P_{PEMA}(\hat{w}|x)$. 
The other pathway directly feeds $r$ into $W_{hd}$ to produce the next-token distribution $P_{LM}(\hat{w}|x)$.
Finally, these two distributions are blended using a tuned parameter $\lambda$ to produce the final distribution of tokens for the desired task as below:
\begin{equation}
    \label{eq:final_distribution}
    \resizebox{.89\hsize}{!}{
    $
    \begin{split}
        P(\hat{w}|x) = \lambda P_{PEMA}(\hat{w}|x) + (1-\lambda) P_{LM}(\hat{w}|x)
    \end{split}
    $
    }
\end{equation}
\section{Gradual Unrolling Interpolation}
\label{s:gu}
\begin{figure}[t]
    \centering
    \includegraphics[width=0.87\linewidth]{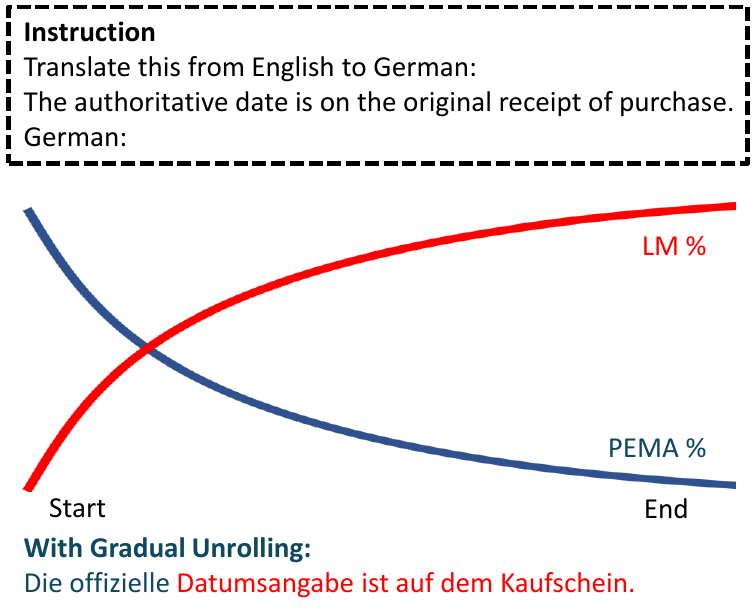}
    \caption{The intuition of Gradual Unrolling. 
    Given the input sentence (Black), the interpolation percentage of the adaptation model (Blue) decreases gradually while that of the language model (Red) increases as the sentence is being generated. This strategy ensures that the adaptation model generates tokens trained for the desired task at the beginning of the sentence, and the language model provides the necessary context in the remaining part of the sentence.}
    \label{fig:gradual_unrolling}
\end{figure}

Given that an adaptation model trained with only a limited number of parameters may lack the context-awareness and language-generation capabilities of pre-trained language models, it is more effective to use the adaptation model to guide the generation of tokens of the desired task at the beginning of the sentence, and rely on a pre-trained language model to provide context for the rest of the sentence. 
To achieve this, we suggest the Gradual Unrolling strategy, which aims for strong $P_{PEMA}(\hat{w}|x)$ interpolation at the beginning of generation and gradually decreases the interpolation. 
As the sentence progresses, the pre-trained language model increasingly contributes to providing the necessary context, as shown in Figure~\ref{fig:gradual_unrolling}.

In the context of sentence generation, we define $SL$ as the input sentence length, excluding instruction and user-defined variables $\lambda_{max}$.
$\lambda$ represents the proportion of the adaptation model's interpolation ($0\leq\lambda\leq1$). 
We also have the dependent variables of the current step ($CS$) and the step size ($SS$). 
The step size is computed as $SS = {\lambda_{max}}/{SL}$, and $CS$ is initialized to $\lambda_{max}$ at the start of sentence generation. 
At each token generation step, $CS$ decreases by $SS$ until the end of the sentence (i.e., $CS_{cur} = CS_{past} - SS$ where $CS_{past}$ is the latest token's $CS$ variable). 
Then, we calculate the current interpolation proportion $\lambda_{cur}$ (i.e., $\lambda$ at Equation~\ref{eq:final_distribution}) as $\lambda_{cur} = CS_{cur}^2$.
\section{Experiments}
This section describes the experiments and results to show both the computational efficiency and performance in downstream tasks of \sys.
First, we perform an experiment on the computational efficiency of \sys.
Subsequently, we evaluate \sys across two downstream tasks: the WMT22 EN$\rightarrow$DE machine translation task~\cite{gpt-mt-2023, kocmi-etal-2022-findings} and the GYAFC formal style transfer task~\cite{gyafc}. 
Lastly, we conduct an ablation study to show the gradual improvement by incorporating each idea of \sys.

\subsection{Computational Efficiency}
To evaluate the computational efficiency of \sys, we conduct a comparison of different fine-tuning methods based on their resource utilization during both training and inference.
We follow the approach of previous work~\cite{pope2023efficiently} that employs a fixed size of input tensors.
We use input tensors with the size [1, 10], equivalent to sequences of 10 tokens with OPT-IML-MAX-1.3B.
The resource utilization metrics encompass training memory consumption, training latency, inference memory consumption, inference latency, and floating point operations per token. 

The evaluation involves several steps. 
First, we clear the CUDA cache to compute the memory and ensure no background GPU processes.
GPU memory utilization is determined using the memory\_summary function provided by Pytorch~\cite{NEURIPS2019_9015}. 
We calculate the time difference before inputting the data into the model and after obtaining the output. For training latency, we consider the time encompassing the entire backpropagation process. To ensure the accuracy of latency, we compute the mean and variance based on ten trials of inputs for each fine-tuning method.
{We conducted a comparative analysis with the offsite-tuning baseline approach, Offsite-Tuning~\cite{xiao2023offsite}. Offsite-Tuning involves knowledge distillation (OT Emulator) and downstream task training using the OT Emulator (OT Plug-in). Subsequently, it utilizes the OT Plug-in to interact with the PLM during the inference phase.}

As shown in Table~\ref{tab:efficiency}, \sys demonstrates the efficiency by utilizing one-tenth of the training memory consumption compared to LoRA. 
In addition, \sys shows the fastest training latency among all the methods. This is because \sys uses external memory to store context representations and does not require access to a pre-trained language model during the training phase, as illustrated in Figure~\ref{fig:model}. These results highlight the significance of \sys's reduced training memory consumption and improved training latency, making it an appealing choice for efficient natural language generation tasks.


\begin{table}[t!]
    \centering
    \resizebox{0.48\textwidth}{!}{
        \begin{tabular}{lrrrrr}
            \toprule

Method & \multicolumn{1}{c}{Tr-MC} & \multicolumn{1}{c}{Tr-Lat} & \multicolumn{1}{c}{Inf-MC} & \multicolumn{1}{c}{Inf-Lat} & \multicolumn{1}{c}{FLOPs}\\
\midrule
FT& 20,082& 250.4$_{\pm140.6}$ & 5,021 & 17.1$_{\pm1.0}$ & 2.41e9\\
FT-top2 & 7,355& 70.3$_{\pm108.6}$ & 5,021 & 17.3$_{\pm1.3}$ & 2.41e9\\
$k$NN-LM & None& 20.3$_{\pm567.2}$ & 5,021  & 37.5$_{\pm1.4}$ & FT+6.29e6\\
LoRA & 5,056& 21.6$_{\pm0.4}$  & 5,031  & 20.5$_{\pm1.5}$ & FT+4.19e6\\
UniPELT & 5,138& 30.3$_{\pm0.1}$  & 5,047 & 21.3$_{\pm0.6}$ & FT+1.49e7\\
OT Emulator & 11,713 & 88.4$_{\pm309.4}$ & None & None & FT+8.03e8 \\
OT Plug-in & 5,267 & 59.6$_{\pm107.8}$ & 5,269 & 21.3$_{\pm0.1}$ & FT+4.82e8 \\
\sys& 478& 18.5$_{\pm1.0}$  & 5,043  & 18.2$_{\pm0.5}$ & FT+4.19e6 \\
\bottomrule
        \end{tabular}
    }
    \caption{Comparison of various training and inference resource utilization methods with OPT-IML-MAX-1.3B.
    We evaluate memory consumption (MC) and latency (Lat) for training (Tr) and inference (Inf), as well as FLOPs per token, using 10-token length sequences.
    Memory size is measured in megabytes, and latency is measured in milliseconds.
    \sys stands out by using only one-tenth of the training memory utilized by LoRA. Furthermore, \sys demonstrates the fastest training latency among the methods.}
    \label{tab:efficiency}
\end{table}
\begin{table*}[t!]
    \centering
    \resizebox{0.97\textwidth}{!}
    {
        \begin{tabular}{lr|rrr|rrr|rrr}
            \toprule
\multirow{2}{*}{Model} & \multicolumn{1}{c}{} & \multicolumn{3}{c}{WMT22 (EN$\rightarrow$DE)}& \multicolumn{3}{c}{GYAFC (F\&R)}  & \multicolumn{3}{c}{GYAFC (E\&M)} \\
& \multicolumn{1}{c}{Tr-MC (MB)}& \multicolumn{1}{c}{sBLEU} & \multicolumn{1}{c}{PPL} & \multicolumn{1}{c}{COMET} & \multicolumn{1}{c}{sBLEU} & \multicolumn{1}{c}{PPL} & \multicolumn{1}{c}{FormImp} & \multicolumn{1}{c}{sBLEU} & \multicolumn{1}{c}{PPL} & \multicolumn{1}{c}{FormImp} \\
\midrule
OPT-1.3B  & None& 9.55 & 51.30 & 57.24& 55.00& \textbf{18.98} & 11.05& 53.98& \underline{20.89} & 10.67\\
OPT-1.3B (FT)& 20,082 & {\underline{10.15}}& \underline{40.83} & {\underline{61.44}}& 29.17& 24.82 & {\underline{52.28}}& 31.50& 27.99 & {46.82}\\
OPT-1.3B (FT-Top2) & 7,355 & 3.57 & 51.36 & 38.35& 21.60& 24.33 & \textbf{59.00}& 23.94& 27.07 & \underline{51.52}\\
OPT-1.3B ($k$NN-LM)  & None& 8.07 & 91.37 & 41.75& 56.69& 20.87 & 16.26&  54.74 & 23.15 & 14.46\\
OPT-1.3B (LoRA) & 5,025 & 4.28 & 61.25 & 39.32& 20.98& \underline{19.07} & 45.71& 15.57& \textbf{19.71} & 46.32 \\
OPT-1.3B (UniPELT) & 5,138 & 9.15 & 47.09 & 56.30 & 51.38 & 44.43 & 52.22& 46.67 & 22.08 & \textbf{53.31}\\
OPT-1.3B (Offsite-Tuning) & 5,267 & 7.65 & \textbf{36.91} & 52.85 & \underline{59.01} & 20.70 & 24.82 & \underline{57.01} & 23.25 & 23.76\\
OPT-1.3B (PEMA) & 478 & \textbf{12.87}& 42.62 & \textbf{64.16}& \textbf{64.82}& 23.15 & 41.90& \textbf{61.24}& 24.28 & 36.28\\
\midrule
LLaMA-7B & None& 2.78 & 78.49 & 39.49& 20.18& 34.53 & 42.81& 24.14& 37.33 & 44.81\\
LLaMA-7B ($k$NN-LM) & None& 0.07 & 85.09 & 38.53& 1.72 & 41.50  & 55.13& 1.94 & 46.31 & {\underline{68.61}}\\
LLaMA-7B (LoRA)& 13,237 & {\underline{11.46}}& \underline{51.36} & {\underline{67.48}}& 52.67& \textbf{22.42} & \textbf{72.23}& {52.15}& \textbf{24.74} & \textbf{71.28}\\

LLaMA-7B (UniPELT)& 13,810 & 9.13 & \textbf{46.62} & 56.31 & \underline{59.81} & \underline{22.95} & \underline{71.69} & \underline{58.07} & \underline{25.35} & 68.33\\

LLaMA-7B (PEMA)& 996 & \textbf{14.50}& 54.26 & \textbf{70.31}& \textbf{63.99}& 23.19 & {61.40}& \textbf{60.88}& 26.00 & 60.94\\
\midrule
OPT-30B&None& {\underline{18.22}}& \textbf{45.81} & {\underline{77.41}}& 60.41& {\textbf{20.04}} & 29.33& 57.60& \textbf{21.97} & 23.88\\
OPT-30B ($k$NN-LM)&None& 16.65& 74.06 & 62.98& 61.02& \underline{20.86} & 29.80& 58.58& \underline{22.75} & 23.39\\
OPT-30B (LoRA)  & 58,083   & 8.26 & 46.97 & 69.41& 61.39& 22.00 & \textbf{73.10}& {\underline{59.76}}& 23.97 & \textbf{68.29}\\
OPT-30B (UniPELT)  & 59,028   & 15.57 & 47.34 & 73.42& \underline{64.54} & 21.72 & 47.14 & 56.86 & 23.77 & 34.08\\
OPT-30B (PEMA)  & 1,909 & \textbf{19.22}& {\underline{46.62}} & \textbf{79.21}& \textbf{70.84}& 22.04 & {\underline{52.35}}& \textbf{65.43}& 25.53 & {\underline{44.63}}  \\
\bottomrule
        \end{tabular}
    }
    \caption{Comparison of various models across different tasks.
    The evaluated tasks include WMT22 (EN$\rightarrow$DE) translation and GYAFC Family \& Relationships (F\&R) and GYAFC Entertainment \& Music (E\&M) style transfer. 
    The models considered for evaluation are OPT-IML-MAX-1.3B, LLaMA-7B, and OPT-IML-MAX-30B, each with specific adaptations and configurations.}
    \label{tab:model_comparison_1}
\end{table*}

\subsection{Performance of Downstream Tasks}
We present a comprehensive analysis of the performance of \sys and baseline models on two downstream tasks: the WMT22 (EN$\rightarrow$DE) translation task and the GYAFC task involving Family \& Relationships and Entertainment \& Music. All tasks are evaluated using zero-shot inference.

For the machine translation task, we use the EN$\rightarrow$DE news-commentary dataset to address the limitation noted in~\cite{brown2020language}, where translations into English tend to be stronger than those from English due to training set biases. We evaluate our model using the latest test set provided by~\cite{gpt-mt-2023, kocmi-etal-2022-findings}.

For the formality style transfer task, we use the GYAFC dataset~\cite{gyafc}, which consists of a parallel training set of informal and formal sentences. The test set comprises four reference sentences paired with one informal sentence.
In this task, our objective is to transfer the style of informal sentences into formal ones.

We use three pre-trained language models: OPT-IML-MAX-1.3B, LLaMA-7B, and OPT-IML-MAX-30B~\cite{iyer2022opt, touvron2023llama}. 
We compare \sys with the following methods:


\begin{itemize}
    \item \textbf{Full fine-tuning (FT)} updates all pre-trained model parameters, including weights and biases.
    \item \textbf{Fine-tuning top-2 (FT-Top2)} updates the last two layers while the remaining layers are frozen.
    \item \textbf{$k$-Nearest Neighbors Language Model ($k$NN-LM)}~\cite{khandelwal20generalization} uses $k$NN search within an external memory to derive a next-token distribution $P_{kNN}$, which is then interpolated with $P_{LM}$ to produce an adapted next-token distribution.
    \item \textbf{LoRA}~\cite{hu2022lora} uses two additional trainable matrices. We apply LoRA at the last layer output projection matrices in the self-attention module.
    \item \noindent\textbf{UniPELT}~\cite{mao2022unipelt} is a state-of-the-art PEFT method that combines Adapter tuning~\cite{pmlr-v97-houlsby19a}, Prefix tuning~\cite{li-liang-2021-prefix}, and LoRA~\cite{hu2022lora} with a gating mechanism to select the optimal approaches. We apply UniPELT at the last layer.
    \item \textbf{Offsite-Tuning}~\cite{xiao2023offsite} is an offsite-tunable method that uses a distilled PLM emulator with an adapter, which includes multiple copies at the PLM's beginning and end. We use four adapter layers for training and inference.
\end{itemize}








We use widely used evaluation metrics to assess the performance of \sys as follows:

\begin{itemize}
    \item \textbf{Sacre-Bleu (sBLEU)}~\cite{sbleu} is a commonly used metric to calculate the n-gram accuracy between the source and target sentences. It evaluates how well the generated sentence preserves the meaning of the reference and captures target domain distribution. We use the implementation from the official repository\footnote{\url{https://github.com/mjpost/sacreBLEU}}. Higher scores are better.
    \item \textbf{Perplexity (PPL)}~\cite{jelinek1977perplexity} is to assess the fluency of generated sentences. We use pre-trained GPT-2 large~\cite{radford2019language} to calculate the exponential of the negative log-likelihood of a current token given the previous context. Lower scores are better.
    \item \textbf{COMET}~\cite{comet} is a neural network-based metric for assessing machine translation quality. It shows a positive correlation with human judgments. We utilize the default, pre-trained COMET model, \footnote{Unbabel/wmt22-comet-da} for the WMT22. Higher scores are better.
    \item \textbf{Formality Improvement (FormImp)} measure formality improvement based on XFORMAL~\cite{briakou-etal-2021-ola}. To measure the formality score of a sentence, we train a BERT-Large~\cite{bert} on an external formality dataset consisting of 4K human-annotated examples~\cite{tacl_a_00083}. We compute the formality score for each formal reference sentence ($FR$), informal input sentence ($II$), and generated sentence ($G$). 
Then, we measure the relative distance using the formula: $\frac{G}{FR-II}\times100$. 
We employ this metric for the GYAFC task.
Higher scores are better.
\end{itemize}


\subsubsection{Results}
For the WMT22 (EN$\rightarrow$DE) translation task, we evaluated sBLEU, PPL, and COMET metrics. As Table~\ref{tab:model_comparison_1} shows, \sys outperforms baselines in sBLEU and COMET. Offsite-Tuninig, LoRA, and UniPELT perform slightly better than a naive pre-trained language model and \sys in terms of PPL. However, they require more memory consumption for training than \sys. Finally, \sys generates more appropriate translated sentences than other baselines for sBLEU with relatively small memory consumption.

For the GYAFC style transfer task, we evaluated sBLEU, PPL, and Formality Improvement (FormImp) metrics. 
As Table~\ref{tab:model_comparison_1} shows, \sys consistently achieves favorable performance. 
\sys shows the highest sBLEU scores, effectively maintaining meaning preservation across different domains and models.
\sys performs slightly better than a naive pre-trained language model and is comparable to other baselines in terms of FormImp.
Furthermore, we observe a trade-off between sBLEU and formality.
These findings support previous observations in the same formality style transfer task with multilingual formality~\cite{xformal}.

\begin{table}[t]
    \centering
    \resizebox{0.99\linewidth}{!}{
        \begin{tabular}{lrrr}
            \toprule
            WMT22 (EN$\rightarrow$DE) & \multicolumn{1}{c}{sBLEU} & \multicolumn{1}{c}{PPL} & \multicolumn{1}{c}{COMET} \\
\midrule
OPT-30B& 18.22& \textbf{45.81} & 77.41\\
OPT-30B+$B_{pd}$ & 18.74& 48.05 & 77.76\\
OPT-30B+$B_{pd}$+$GU$& \underline{19.17}& 48.60 & \underline{78.57}\\
OPT-30B+$B_{pd}$+$GU$+$B_{rct}$ (\sys)& \textbf{19.22}&\underline{46.62} & \textbf{79.21}\\
\midrule
GYAFC (F\&R)  & \multicolumn{1}{c}{sBLEU} & \multicolumn{1}{c}{PPL} & \multicolumn{1}{c}{FormImp}\\
\midrule
OPT-30B& 60.41& \underline{20.04} & 29.33 \\
OPT-30B+$B_{pd}$ & 70.00& 20.38 & 47.38 \\
OPT-30B+$B_{pd}$+$GU$& \underline{70.29}& \textbf{16.95}  & \underline{51.24} \\
OPT-30B+$B_{pd}$+$GU$+$B_{rct}$ (\sys) & \textbf{70.84} & 22.04  & \textbf{52.35}\\
\midrule
GYAFC (E\&M)  & \multicolumn{1}{c}{sBLEU} & \multicolumn{1}{c}{PPL} & \multicolumn{1}{c}{FormImp}\\
\midrule
OPT-30B& 57.60& \textbf{21.97} & 23.88\\
OPT-30B+$B_{pd}$ & 64.37& 26.76 & 38.80\\
OPT-30B+$B_{pd}$+$GU$& \underline{64.82}& 25.62& \underline{42.61}\\
OPT-30B+$B_{pd}$+$GU$+$B_{rct}$ (\sys)& \textbf{65.43} & \underline{25.53} & \textbf{44.63} \\
\bottomrule
        \end{tabular}
    }
    \caption{Ablation results of \sys over our proposed approaches. The techniques include a token prediction decoder ($B_{pd}$), Gradual Unrolling ($GU$), and a reconstruction decoder ($B_{rct}$). We use OPT-IML-MAX-30B as a baseline. Implementing all techniques together enhances overall performance.}
    \label{tab:ablation}
\end{table}
\begin{table}[t!]
    \centering
    \resizebox{0.49\textwidth}{!}{
        \begin{tabular}{lrrrrrrrrrr}
            \toprule
            $\lambda/\lambda_{max}$  & \multicolumn{1}{c}{0.9} & \multicolumn{1}{c}{0.8} & \multicolumn{1}{c}{0.7} & \multicolumn{1}{c}{0.6} & \multicolumn{1}{c}{0.5} & \multicolumn{1}{c}{0.4} & \multicolumn{1}{c}{0.3} & \multicolumn{1}{c}{0.2} & \multicolumn{1}{c}{0.1} \\
\midrule
With $GU$ & 47.45 & 46.61 & 46.62 & 46.18 & 46.12 & 46.03 & 45.85 & 45.89 & 45.84 \\
Without $GU$ & 54.29 & 51.87 & 50.22 & 49.70 & 49.45 & 48.09 & 47.76 & 47.67 & 47.52 \\
\bottomrule
        \end{tabular}
    }
    \caption{
    Impact of Gradual Unrolling ($GU$) on perplexity across different $\lambda/\lambda_{max}$ values.
    Using $GU$ consistently outperforms the approach without $GU$ for all $\lambda/\lambda_{max}$ values, ranging from $0.1$ to $0.9$.}
    \label{tab:lambda_ppl}
\end{table}

\begin{figure}
    \centering
    \includegraphics[width=0.97\linewidth]{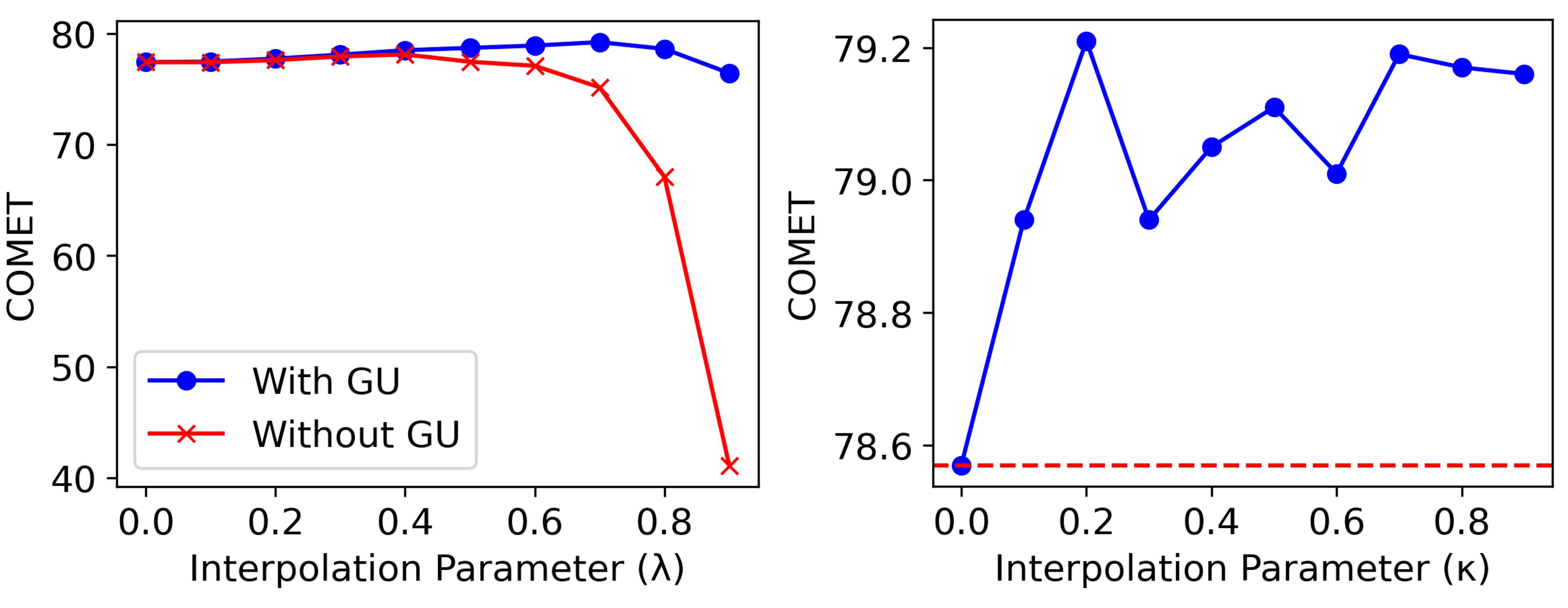}
    \caption{Performance variations on the WMT22 task with interpolation values $\lambda_{max}$ (left) and $\kappa$ (right). For $\lambda_{max}$, using Gradual Unrolling ($GU$) prevents performance degradation and enhances results, unlike without $GU$, where performance drops sharply. With $\kappa$ when $\lambda_{max}$ is set at 0.7, combining reconstruction loss with next-token prediction loss improves performance over excluding reconstruction loss (red dotted line), as indicated by better results when $\kappa$ is above zero.}
    \label{fig:lambda_kappa}
\end{figure}


\subsection{Ablation Study}
To assess the effectiveness of \sys, we conduct ablation studies to demonstrate the incremental improvement achieved by incorporating each component of \sys. We utilize a token prediction decoder ($B_{pd}$) to predict the target token based on the context representation obtained from the pre-trained language model. As shown in Table~\ref{tab:ablation}, the token prediction decoder enhances task performance. Building on this, we incorporated Gradual Unrolling ($GU$) and the Reconstruction Decoder ($B_{rct}$) to further improve performance. The inclusion of these three methods yields the highest performance gains, as shown in the results.


\noindent
\textbf{Interpolation Parameter ($\lambda_{max}$)}
We propose the Gradual Unrolling (GU) interpolation strategy, where \sys initially guides the generation of a new task and subsequently leverages the language model for contextual completion of sentences.
Table~\ref{tab:ablation} shows the effectiveness of $GU$ in enhancing performance by enabling the language model to provide context completion.
We further compare with and without $GU$ by adjusting the $\lambda_{max}$ hyperparameter in the WMT22 task.
As shown in Figure~\ref{fig:lambda_kappa}, with $GU$ maintains better performance stability at higher $\lambda_{max}$ values while achieving noticeable performance improvement over without $GU$. We also report details on the impact of incorporating $\lambda_{max}$ in Figure~\ref{fig:lambda} in the appendix.
Additionally, we conduct an experiment to measure perplexity.
Table~\ref{tab:lambda_ppl} shows that $GU$ consistently outperforms across $\lambda/\lambda_{max}$ values from 0.1 to 0.9.

\noindent
\textbf{Interpolation Parameter ($\kappa$)}
We investigate the effectiveness of the reconstruction decoder, which reconstructs the original vector $f(c_i)$. 
Table~\ref{tab:ablation} and Figure~\ref{fig:lambda_kappa} demonstrate that incorporating the reconstruction decoder improves performance across desired tasks, demonstrating its efficacy in enhancing generation quality. We also report details on the impact of incorporating $\kappa$ in Figure~\ref{fig:kappa_wmt} in the appendix.

\section{Conclusion}
In this paper, we present \sys, a novel parameter-efficient fine-tuning approach for language models. Unlike existing PEFT methods, 
{\sys utilizes minimal pre-trained model parameters during training, making it an efficient and adaptable method for offsite-tuning.}
\sys includes a token prediction decoder, Gradual Unrolling, and a reconstruction decoder to improve model performance.
Our comprehensive evaluations on translation and style transfer tasks demonstrate \sys's effectiveness in generating text that more closely follows target domain distributions.
Additionally, \sys proves its computational efficiency by utilizing minimal training memory and achieving faster training latency with a syntactic dataset.
Overall, \sys offers efficient fine-tuning and presents a promising direction for an offsite-tunable PEFT approach in downstream NLP tasks.

\section*{Limitations and Future Work}
\noindent
\textbf{Privacy Concern at Inference} \sys introduces a novel Parameter-Efficient Fine-Tuning (PEFT) method for privacy-preserving offsite-tuning. However, this process requires data owners to share predicted next-tokens with PLM owners during inference, which raises potential privacy concerns. These concerns necessitate further investigation of effective mitigation strategies.

\noindent
\textbf{Shared PLM Weight with Data Owner} Sharing the $W_{hd}$ weight between PLM owners and data owners poses challenges related to model privacy. In our experiments, we used open-source PLMs due to the confidentiality issues associated with proprietary PLMs. Our future work will explore enabling data owners to generate a new Language Model (LM) head using a shared tokenizer from the PLM owner, enhancing privacy between the PLM and the data owner.

\noindent
\textbf{Unintentional Data Leakage} Through PEMA, data and PLM owners can fine-tune efficiently and effectively with minimal communication. However, how data owners use \sys could unintentionally lead to data leakage issues. Subsequent research will explore solutions to address this challenge.

\noindent
\textbf{Other Applications}
While our research has been focused on machine translation tasks, it can be applied to various NLP tasks depending on the initial input. Consequently, future studies will investigate the application of our method across a range of NLP tasks.

\noindent
\textbf{Practical Applicability}
\sys provides offsite-tuning under conditions of limited information sharing, specifically the context representation, LM head, and next-token probability from PLMs. We achieve this by learning downstream tasks using features similar to those accessible from OpenAI API, such as Embedding API\footnote{\url{https://platform.openai.com/docs/guides/embeddings/what-are-embeddings}}, which we relate to context representation and next-token probability\footnote{\url{https://platform.openai.com/docs/api-reference/chat/create\#chat-create-logprobs}}. However, our current setup does not extend to practical implementation of fine-tuning current proprietary PLMs (e.g., OpenAI, Claude) fully. The primary issue is current proprietary PLMs do not share LM head.

Nevertheless, some proprietary LLMs, such as OpenAI, share their tokenizer publicly\footnote{\url{https://platform.openai.com/tokenizer}}. This tokenizer shows the method of text splitting and provides token indexes. We believe the availability of tokenizers will be beneficial for future research in overcoming limitations related to sharing the LM head. For a more detailed explanation, LM Head ($\mathbb{R}^{v \times d}$) outputs the probability of each token index. Through $B_{pd}$ of \sys, $h_{pd}$ ($\mathbb{R}^{d\times 1}$) is created. Therefore, it is possible to directly predict the probability of a token using a separate LM head that outputs ($\mathbb{R}^{v\times 1}$) if we know $v$ and the index of each token. We posit that access to tokenizers offers an opportunity for data owners to construct a new, distinct LM head compatible with \sys.

\section*{Ethics Statement}
The results of our research are based on existing studies, and all generation models and datasets used are publicly available and used for their intended use with no ethical concerns.
\section*{Acknowledgements}
We would like to thank the anonymous reviewers for their helpful questions and comments.
This research was partly supported by the Bio \& Medical Technology Development Program of the National Research Foundation (NRF) funded by the Korean government (MSIT) (NRF-2021M3A9E4080780),  
Institute of Information \& communications Technology Planning \& Evaluation (IITP) grant funded by the Korea government (MSIT) (No.2019-0-00421, AI Graduate School Support Program(Sungkyunkwan University) and IITP-2023-2020-0-018, abductive inference framework using omni-data for understanding complex causal relations \& ICT Creative Consilience program).
\bibliography{main2}

\begin{thebibliography}{47}
\expandafter\ifx\csname natexlab\endcsname\relax\def\natexlab#1{#1}\fi

\bibitem[{Anil et~al.(2023)Anil, Dai, Firat, Johnson, Lepikhin, Passos, Shakeri, Taropa, Bailey, Chen, Chu, Clark, Shafey, Huang, Meier-Hellstern, Mishra, Moreira, Omernick, Robinson, Ruder, Tay, Xiao, Xu, Zhang, Abrego, Ahn, Austin, Barham, Botha, Bradbury, Brahma, Brooks, Catasta, Cheng, Cherry, Choquette-Choo, Chowdhery, Crepy, Dave, Dehghani, Dev, Devlin, Díaz, Du, Dyer, Feinberg, Feng, Fienber, Freitag, Garcia, Gehrmann, Gonzalez, Gur-Ari, Hand, Hashemi, Hou, Howland, Hu, Hui, Hurwitz, Isard, Ittycheriah, Jagielski, Jia, Kenealy, Krikun, Kudugunta, Lan, Lee, Lee, Li, Li, Li, Li, Li, Lim, Lin, Liu, Liu, Maggioni, Mahendru, Maynez, Misra, Moussalem, Nado, Nham, Ni, Nystrom, Parrish, Pellat, Polacek, Polozov, Pope, Qiao, Reif, Richter, Riley, Ros, Roy, Saeta, Samuel, Shelby, Slone, Smilkov, So, Sohn, Tokumine, Valter, Vasudevan, Vodrahalli, Wang, Wang, Wang, Wang, Wieting, Wu, Xu, Xu, Xue, Yin, Yu, Zhang, Zheng, Zheng, Zhou, Zhou, Petrov, and Wu}]{anil2023palm}
Rohan Anil, Andrew~M. Dai, Orhan Firat, Melvin Johnson, Dmitry Lepikhin, Alexandre Passos, Siamak Shakeri, Emanuel Taropa, Paige Bailey, Zhifeng Chen, Eric Chu, Jonathan~H. Clark, Laurent~El Shafey, Yanping Huang, Kathy Meier-Hellstern, Gaurav Mishra, Erica Moreira, Mark Omernick, Kevin Robinson, Sebastian Ruder, Yi~Tay, Kefan Xiao, Yuanzhong Xu, Yujing Zhang, Gustavo~Hernandez Abrego, Junwhan Ahn, Jacob Austin, Paul Barham, Jan Botha, James Bradbury, Siddhartha Brahma, Kevin Brooks, Michele Catasta, Yong Cheng, Colin Cherry, Christopher~A. Choquette-Choo, Aakanksha Chowdhery, Clément Crepy, Shachi Dave, Mostafa Dehghani, Sunipa Dev, Jacob Devlin, Mark Díaz, Nan Du, Ethan Dyer, Vlad Feinberg, Fangxiaoyu Feng, Vlad Fienber, Markus Freitag, Xavier Garcia, Sebastian Gehrmann, Lucas Gonzalez, Guy Gur-Ari, Steven Hand, Hadi Hashemi, Le~Hou, Joshua Howland, Andrea Hu, Jeffrey Hui, Jeremy Hurwitz, Michael Isard, Abe Ittycheriah, Matthew Jagielski, Wenhao Jia, Kathleen Kenealy, Maxim Krikun, Sneha Kudugunta, Chang
  Lan, Katherine Lee, Benjamin Lee, Eric Li, Music Li, Wei Li, YaGuang Li, Jian Li, Hyeontaek Lim, Hanzhao Lin, Zhongtao Liu, Frederick Liu, Marcello Maggioni, Aroma Mahendru, Joshua Maynez, Vedant Misra, Maysam Moussalem, Zachary Nado, John Nham, Eric Ni, Andrew Nystrom, Alicia Parrish, Marie Pellat, Martin Polacek, Alex Polozov, Reiner Pope, Siyuan Qiao, Emily Reif, Bryan Richter, Parker Riley, Alex~Castro Ros, Aurko Roy, Brennan Saeta, Rajkumar Samuel, Renee Shelby, Ambrose Slone, Daniel Smilkov, David~R. So, Daniel Sohn, Simon Tokumine, Dasha Valter, Vijay Vasudevan, Kiran Vodrahalli, Xuezhi Wang, Pidong Wang, Zirui Wang, Tao Wang, John Wieting, Yuhuai Wu, Kelvin Xu, Yunhan Xu, Linting Xue, Pengcheng Yin, Jiahui Yu, Qiao Zhang, Steven Zheng, Ce~Zheng, Weikang Zhou, Denny Zhou, Slav Petrov, and Yonghui Wu. 2023.
\newblock \href {http://arxiv.org/abs/2305.10403} {Palm 2 technical report}.

\bibitem[{AnthropicAI(2023)}]{claude}
AnthropicAI. 2023.
\newblock Introducing claude.
\newblock \url{https://www.anthropic.com/index/introducing-claude}.
\newblock Accessed: 2023-08-15.

\bibitem[{Briakou et~al.(2021{\natexlab{a}})Briakou, Lu, Zhang, and Tetreault}]{briakou-etal-2021-ola}
Eleftheria Briakou, Di~Lu, Ke~Zhang, and Joel Tetreault. 2021{\natexlab{a}}.
\newblock \href {https://doi.org/10.18653/v1/2021.naacl-main.256} {Ol{\'a}, bonjour, salve! {XFORMAL}: A benchmark for multilingual formality style transfer}.
\newblock In \emph{Proceedings of the 2021 Conference of the North American Chapter of the Association for Computational Linguistics: Human Language Technologies}, pages 3199--3216, Online. Association for Computational Linguistics.

\bibitem[{Briakou et~al.(2021{\natexlab{b}})Briakou, Lu, Zhang, and Tetreault}]{xformal}
Eleftheria Briakou, Di~Lu, Ke~Zhang, and Joel Tetreault. 2021{\natexlab{b}}.
\newblock \href {https://doi.org/10.18653/v1/2021.naacl-main.256} {Ol{\'a}, bonjour, salve! {XFORMAL}: A benchmark for multilingual formality style transfer}.
\newblock In \emph{Proceedings of the 2021 Conference of the North American Chapter of the Association for Computational Linguistics: Human Language Technologies}, pages 3199--3216, Online. Association for Computational Linguistics.

\bibitem[{Brown et~al.(2020)Brown, Mann, Ryder, Subbiah, Kaplan, Dhariwal, Neelakantan, Shyam, Sastry, Askell, Agarwal, Herbert-Voss, Krueger, Henighan, Child, Ramesh, Ziegler, Wu, Winter, Hesse, Chen, Sigler, Litwin, Gray, Chess, Clark, Berner, McCandlish, Radford, Sutskever, and Amodei}]{brown2020language}
Tom Brown, Benjamin Mann, Nick Ryder, Melanie Subbiah, Jared~D Kaplan, Prafulla Dhariwal, Arvind Neelakantan, Pranav Shyam, Girish Sastry, Amanda Askell, Sandhini Agarwal, Ariel Herbert-Voss, Gretchen Krueger, Tom Henighan, Rewon Child, Aditya Ramesh, Daniel Ziegler, Jeffrey Wu, Clemens Winter, Chris Hesse, Mark Chen, Eric Sigler, Mateusz Litwin, Scott Gray, Benjamin Chess, Jack Clark, Christopher Berner, Sam McCandlish, Alec Radford, Ilya Sutskever, and Dario Amodei. 2020.
\newblock \href {https://proceedings.neurips.cc/paper_files/paper/2020/file/1457c0d6bfcb4967418bfb8ac142f64a-Paper.pdf} {Language models are few-shot learners}.
\newblock In \emph{Advances in Neural Information Processing Systems}, volume~33, pages 1877--1901. Curran Associates, Inc.

\bibitem[{Chowdhery et~al.(2023)Chowdhery, Narang, Devlin, Bosma, Mishra, Roberts, Barham, Chung, Sutton, Gehrmann, Schuh, Shi, Tsvyashchenko, Maynez, Rao, Barnes, Tay, Shazeer, Prabhakaran, Reif, Du, Hutchinson, Pope, Bradbury, Austin, Isard, Gur-Ari, Yin, Duke, Levskaya, Ghemawat, Dev, Michalewski, Garcia, Misra, Robinson, Fedus, Zhou, Ippolito, Luan, Lim, Zoph, Spiridonov, Sepassi, Dohan, Agrawal, Omernick, Dai, Pillai, Pellat, Lewkowycz, Moreira, Child, Polozov, Lee, Zhou, Wang, Saeta, Diaz, Firat, Catasta, Wei, Meier-Hellstern, Eck, Dean, Petrov, and Fiedel}]{chowdhery2022palm}
Aakanksha Chowdhery, Sharan Narang, Jacob Devlin, Maarten Bosma, Gaurav Mishra, Adam Roberts, Paul Barham, Hyung~Won Chung, Charles Sutton, Sebastian Gehrmann, Parker Schuh, Kensen Shi, Sasha Tsvyashchenko, Joshua Maynez, Abhishek Rao, Parker Barnes, Yi~Tay, Noam Shazeer, Vinodkumar Prabhakaran, Emily Reif, Nan Du, Ben Hutchinson, Reiner Pope, James Bradbury, Jacob Austin, Michael Isard, Guy Gur-Ari, Pengcheng Yin, Toju Duke, Anselm Levskaya, Sanjay Ghemawat, Sunipa Dev, Henryk Michalewski, Xavier Garcia, Vedant Misra, Kevin Robinson, Liam Fedus, Denny Zhou, Daphne Ippolito, David Luan, Hyeontaek Lim, Barret Zoph, Alexander Spiridonov, Ryan Sepassi, David Dohan, Shivani Agrawal, Mark Omernick, Andrew~M. Dai, Thanumalayan~Sankaranarayana Pillai, Marie Pellat, Aitor Lewkowycz, Erica Moreira, Rewon Child, Oleksandr Polozov, Katherine Lee, Zongwei Zhou, Xuezhi Wang, Brennan Saeta, Mark Diaz, Orhan Firat, Michele Catasta, Jason Wei, Kathy Meier-Hellstern, Douglas Eck, Jeff Dean, Slav Petrov, and Noah Fiedel. 2023.
\newblock \href {http://jmlr.org/papers/v24/22-1144.html} {Palm: Scaling language modeling with pathways}.
\newblock \emph{Journal of Machine Learning Research}, 24(240):1--113.

\bibitem[{Devlin et~al.(2019)Devlin, Chang, Lee, and Toutanova}]{bert}
Jacob Devlin, Ming-Wei Chang, Kenton Lee, and Kristina Toutanova. 2019.
\newblock \href {https://doi.org/10.18653/v1/N19-1423} {{BERT}: Pre-training of deep bidirectional transformers for language understanding}.
\newblock In \emph{Proceedings of the 2019 Conference of the North {A}merican Chapter of the Association for Computational Linguistics: Human Language Technologies, Volume 1 (Long and Short Papers)}, pages 4171--4186, Minneapolis, Minnesota. Association for Computational Linguistics.

\bibitem[{Ding et~al.(2022)Ding, Qin, Yang, Wei, Yang, Su, Hu, Chen, Chan, Chen, Yi, Zhao, Wang, Liu, Zheng, Chen, Liu, Tang, Li, and Sun}]{ding2022delta}
Ning Ding, Yujia Qin, Guang Yang, Fuchao Wei, Zonghan Yang, Yusheng Su, Shengding Hu, Yulin Chen, Chi-Min Chan, Weize Chen, Jing Yi, Weilin Zhao, Xiaozhi Wang, Zhiyuan Liu, Hai-Tao Zheng, Jianfei Chen, Yang Liu, Jie Tang, Juanzi Li, and Maosong Sun. 2022.
\newblock \href {http://arxiv.org/abs/2203.06904} {Delta tuning: A comprehensive study of parameter efficient methods for pre-trained language models}.

\bibitem[{Guinney and Saez-Rodriguez(2018)}]{guinney2018alternative}
Justin Guinney and Julio Saez-Rodriguez. 2018.
\newblock \href {https://doi.org/https://doi.org/10.1038/nbt.4128} {Alternative models for sharing confidential biomedical data}.
\newblock \emph{Nature biotechnology}, 36(5):391--392.

\bibitem[{He et~al.(2022)He, Zhou, Ma, Berg-Kirkpatrick, and Neubig}]{he2021towards}
Junxian He, Chunting Zhou, Xuezhe Ma, Taylor Berg-Kirkpatrick, and Graham Neubig. 2022.
\newblock \href {https://openreview.net/forum?id=0RDcd5Axok} {Towards a unified view of parameter-efficient transfer learning}.
\newblock In \emph{International Conference on Learning Representations}.

\bibitem[{Hendy et~al.(2023)Hendy, Abdelrehim, Sharaf, Raunak, Gabr, Matsushita, Kim, Afify, and Awadalla}]{gpt-mt-2023}
Amr Hendy, Mohamed Abdelrehim, Amr Sharaf, Vikas Raunak, Mohamed Gabr, Hitokazu Matsushita, Young~Jin Kim, Mohamed Afify, and Hany~Hassan Awadalla. 2023.
\newblock \href {http://arxiv.org/abs/2302.09210} {How good are gpt models at machine translation? a comprehensive evaluation}.

\bibitem[{Houlsby et~al.(2019)Houlsby, Giurgiu, Jastrzebski, Morrone, De~Laroussilhe, Gesmundo, Attariyan, and Gelly}]{pmlr-v97-houlsby19a}
Neil Houlsby, Andrei Giurgiu, Stanislaw Jastrzebski, Bruna Morrone, Quentin De~Laroussilhe, Andrea Gesmundo, Mona Attariyan, and Sylvain Gelly. 2019.
\newblock \href {https://proceedings.mlr.press/v97/houlsby19a.html} {Parameter-efficient transfer learning for {NLP}}.
\newblock In \emph{Proceedings of the 36th International Conference on Machine Learning}, volume~97 of \emph{Proceedings of Machine Learning Research}, pages 2790--2799. PMLR.

\bibitem[{Hu et~al.(2022)Hu, Shen, Wallis, Allen-Zhu, Li, Wang, Wang, and Chen}]{hu2022lora}
Edward~J Hu, Yelong Shen, Phillip Wallis, Zeyuan Allen-Zhu, Yuanzhi Li, Shean Wang, Lu~Wang, and Weizhu Chen. 2022.
\newblock \href {https://openreview.net/forum?id=nZeVKeeFYf9} {Lo{RA}: Low-rank adaptation of large language models}.
\newblock In \emph{International Conference on Learning Representations}.

\bibitem[{Huang et~al.(2020)Huang, Li, and Yao}]{8733017}
Changwu Huang, Yuanxiang Li, and Xin Yao. 2020.
\newblock \href {https://doi.org/10.1109/TEVC.2019.2921598} {A survey of automatic parameter tuning methods for metaheuristics}.
\newblock \emph{IEEE Transactions on Evolutionary Computation}, 24(2):201--216.

\bibitem[{Iyer et~al.(2023)Iyer, Lin, Pasunuru, Mihaylov, Simig, Yu, Shuster, Wang, Liu, Koura, Li, O'Horo, Pereyra, Wang, Dewan, Celikyilmaz, Zettlemoyer, and Stoyanov}]{iyer2022opt}
Srinivasan Iyer, Xi~Victoria Lin, Ramakanth Pasunuru, Todor Mihaylov, Daniel Simig, Ping Yu, Kurt Shuster, Tianlu Wang, Qing Liu, Punit~Singh Koura, Xian Li, Brian O'Horo, Gabriel Pereyra, Jeff Wang, Christopher Dewan, Asli Celikyilmaz, Luke Zettlemoyer, and Ves Stoyanov. 2023.
\newblock \href {http://arxiv.org/abs/2212.12017} {Opt-iml: Scaling language model instruction meta learning through the lens of generalization}.

\bibitem[{Jelinek et~al.(2005)Jelinek, Mercer, Bahl, and Baker}]{jelinek1977perplexity}
F.~Jelinek, R.~L. Mercer, L.~R. Bahl, and J.~K. Baker. 2005.
\newblock \href {https://doi.org/10.1121/1.2016299} {{Perplexity—a measure of the difficulty of speech recognition tasks}}.
\newblock \emph{The Journal of the Acoustical Society of America}, 62(S1):S63--S63.

\bibitem[{Jiang et~al.(2024)Jiang, Sablayrolles, Roux, Mensch, Savary, Bamford, Chaplot, de~las Casas, Hanna, Bressand, Lengyel, Bour, Lample, Lavaud, Saulnier, Lachaux, Stock, Subramanian, Yang, Antoniak, Scao, Gervet, Lavril, Wang, Lacroix, and Sayed}]{jiang2024mixtral}
Albert~Q. Jiang, Alexandre Sablayrolles, Antoine Roux, Arthur Mensch, Blanche Savary, Chris Bamford, Devendra~Singh Chaplot, Diego de~las Casas, Emma~Bou Hanna, Florian Bressand, Gianna Lengyel, Guillaume Bour, Guillaume Lample, Lélio~Renard Lavaud, Lucile Saulnier, Marie-Anne Lachaux, Pierre Stock, Sandeep Subramanian, Sophia Yang, Szymon Antoniak, Teven~Le Scao, Théophile Gervet, Thibaut Lavril, Thomas Wang, Timothée Lacroix, and William~El Sayed. 2024.
\newblock \href {http://arxiv.org/abs/2401.04088} {Mixtral of experts}.

\bibitem[{Khandelwal et~al.(2021)Khandelwal, Fan, Jurafsky, Zettlemoyer, and Lewis}]{khandelwal2021nearest}
Urvashi Khandelwal, Angela Fan, Dan Jurafsky, Luke Zettlemoyer, and Mike Lewis. 2021.
\newblock \href {https://openreview.net/forum?id=7wCBOfJ8hJM} {Nearest neighbor machine translation}.
\newblock In \emph{International Conference on Learning Representations}.

\bibitem[{Khandelwal et~al.(2020)Khandelwal, Levy, Jurafsky, Zettlemoyer, and Lewis}]{khandelwal20generalization}
Urvashi Khandelwal, Omer Levy, Dan Jurafsky, Luke Zettlemoyer, and Mike Lewis. 2020.
\newblock \href {https://openreview.net/forum?id=HklBjCEKvH} {Generalization through memorization: Nearest neighbor language models}.
\newblock In \emph{International Conference on Learning Representations}.

\bibitem[{Kocmi et~al.(2022)Kocmi, Bawden, Bojar, Dvorkovich, Federmann, Fishel, Gowda, Graham, Grundkiewicz, Haddow, Knowles, Koehn, Monz, Morishita, Nagata, Nakazawa, Nov{\'a}k, Popel, and Popovi{\'c}}]{kocmi-etal-2022-findings}
Tom Kocmi, Rachel Bawden, Ond{\v{r}}ej Bojar, Anton Dvorkovich, Christian Federmann, Mark Fishel, Thamme Gowda, Yvette Graham, Roman Grundkiewicz, Barry Haddow, Rebecca Knowles, Philipp Koehn, Christof Monz, Makoto Morishita, Masaaki Nagata, Toshiaki Nakazawa, Michal Nov{\'a}k, Martin Popel, and Maja Popovi{\'c}. 2022.
\newblock \href {https://aclanthology.org/2022.wmt-1.1} {Findings of the 2022 conference on machine translation ({WMT}22)}.
\newblock In \emph{Proceedings of the Seventh Conference on Machine Translation (WMT)}, pages 1--45, Abu Dhabi, United Arab Emirates (Hybrid). Association for Computational Linguistics.

\bibitem[{Lester et~al.(2021)Lester, Al-Rfou, and Constant}]{lester-etal-2021-power}
Brian Lester, Rami Al-Rfou, and Noah Constant. 2021.
\newblock \href {https://doi.org/10.18653/v1/2021.emnlp-main.243} {The power of scale for parameter-efficient prompt tuning}.
\newblock In \emph{Proceedings of the 2021 Conference on Empirical Methods in Natural Language Processing}, pages 3045--3059, Online and Punta Cana, Dominican Republic. Association for Computational Linguistics.

\bibitem[{Li and Liang(2021)}]{li-liang-2021-prefix}
Xiang~Lisa Li and Percy Liang. 2021.
\newblock \href {https://doi.org/10.18653/v1/2021.acl-long.353} {Prefix-tuning: Optimizing continuous prompts for generation}.
\newblock In \emph{Proceedings of the 59th Annual Meeting of the Association for Computational Linguistics and the 11th International Joint Conference on Natural Language Processing (Volume 1: Long Papers)}, pages 4582--4597, Online. Association for Computational Linguistics.

\bibitem[{Liu et~al.(2022)Liu, Ji, Fu, Tam, Du, Yang, and Tang}]{liu2021p}
Xiao Liu, Kaixuan Ji, Yicheng Fu, Weng Tam, Zhengxiao Du, Zhilin Yang, and Jie Tang. 2022.
\newblock \href {https://doi.org/10.18653/v1/2022.acl-short.8} {{P}-tuning: Prompt tuning can be comparable to fine-tuning across scales and tasks}.
\newblock In \emph{Proceedings of the 60th Annual Meeting of the Association for Computational Linguistics (Volume 2: Short Papers)}, pages 61--68, Dublin, Ireland. Association for Computational Linguistics.

\bibitem[{Loshchilov and Hutter(2019)}]{loshchilov2018decoupled}
Ilya Loshchilov and Frank Hutter. 2019.
\newblock \href {https://openreview.net/forum?id=Bkg6RiCqY7} {Decoupled weight decay regularization}.
\newblock In \emph{International Conference on Learning Representations}.

\bibitem[{Mao et~al.(2022)Mao, Mathias, Hou, Almahairi, Ma, Han, Yih, and Khabsa}]{mao2022unipelt}
Yuning Mao, Lambert Mathias, Rui Hou, Amjad Almahairi, Hao Ma, Jiawei Han, Scott Yih, and Madian Khabsa. 2022.
\newblock \href {https://doi.org/10.18653/v1/2022.acl-long.433} {{U}ni{PELT}: A unified framework for parameter-efficient language model tuning}.
\newblock In \emph{Proceedings of the 60th Annual Meeting of the Association for Computational Linguistics (Volume 1: Long Papers)}, pages 6253--6264, Dublin, Ireland. Association for Computational Linguistics.

\bibitem[{OpenAI(2022)}]{chatgpt}
OpenAI. 2022.
\newblock Chatgpt: Optimizing language models for dialogue.
\newblock \url{https://online-chatgpt.com/}.
\newblock Accessed: 2023-08-15.

\bibitem[{OpenAI(2023{\natexlab{a}})}]{openaift}
OpenAI. 2023{\natexlab{a}}.
\newblock Fine-tuning - openai api.
\newblock \url{https://platform.openai.com/docs/guides/fine-tuning}.
\newblock Accessed: 2023-08-15.

\bibitem[{OpenAI(2023{\natexlab{b}})}]{openai2023gpt4}
OpenAI. 2023{\natexlab{b}}.
\newblock \href {http://arxiv.org/abs/2303.08774} {Gpt-4 technical report}.

\bibitem[{Paszke et~al.(2019)Paszke, Gross, Massa, Lerer, Bradbury, Chanan, Killeen, Lin, Gimelshein, Antiga, Desmaison, Kopf, Yang, DeVito, Raison, Tejani, Chilamkurthy, Steiner, Fang, Bai, and Chintala}]{NEURIPS2019_9015}
Adam Paszke, Sam Gross, Francisco Massa, Adam Lerer, James Bradbury, Gregory Chanan, Trevor Killeen, Zeming Lin, Natalia Gimelshein, Luca Antiga, Alban Desmaison, Andreas Kopf, Edward Yang, Zachary DeVito, Martin Raison, Alykhan Tejani, Sasank Chilamkurthy, Benoit Steiner, Lu~Fang, Junjie Bai, and Soumith Chintala. 2019.
\newblock \href {http://papers.neurips.cc/paper/9015-pytorch-an-imperative-style-high-performance-deep-learning-library.pdf} {Pytorch: An imperative style, high-performance deep learning library}.
\newblock In \emph{Advances in Neural Information Processing Systems 32}, pages 8024--8035. Curran Associates, Inc.

\bibitem[{Pavlick and Tetreault(2016)}]{tacl_a_00083}
Ellie Pavlick and Joel Tetreault. 2016.
\newblock \href {https://doi.org/10.1162/tacl_a_00083} {{An Empirical Analysis of Formality in Online Communication}}.
\newblock \emph{Transactions of the Association for Computational Linguistics}, 4:61--74.

\bibitem[{Pestov(2013)}]{pestov2013k}
Vladimir Pestov. 2013.
\newblock \href {https://doi.org/https://doi.org/10.1016/j.camwa.2012.09.011} {Is the k-nn classifier in high dimensions affected by the curse of dimensionality?}
\newblock \emph{Computers \& Mathematics with Applications}, 65(10):1427--1437.
\newblock Grasping Complexity.

\bibitem[{Pfeiffer et~al.(2021)Pfeiffer, Kamath, R{\"u}ckl{\'e}, Cho, and Gurevych}]{pfeiffer-etal-2021-adapterfusion}
Jonas Pfeiffer, Aishwarya Kamath, Andreas R{\"u}ckl{\'e}, Kyunghyun Cho, and Iryna Gurevych. 2021.
\newblock \href {https://doi.org/10.18653/v1/2021.eacl-main.39} {{A}dapter{F}usion: Non-destructive task composition for transfer learning}.
\newblock In \emph{Proceedings of the 16th Conference of the European Chapter of the Association for Computational Linguistics: Main Volume}, pages 487--503, Online. Association for Computational Linguistics.

\bibitem[{Pichai(2023)}]{bard}
Sundar Pichai. 2023.
\newblock An important next step on our ai journey.
\newblock \url{https://blog.google/intl/en-africa/products/explore-get-answers/an-important-next-step-on-our-ai-journey/}.
\newblock Accessed: 2023-08-15.

\bibitem[{Pope et~al.(2023)Pope, Douglas, Chowdhery, Devlin, Bradbury, Heek, Xiao, Agrawal, and Dean}]{pope2023efficiently}
Reiner Pope, Sholto Douglas, Aakanksha Chowdhery, Jacob Devlin, James Bradbury, Jonathan Heek, Kefan Xiao, Shivani Agrawal, and Jeff Dean. 2023.
\newblock \href {https://proceedings.mlsys.org/paper_files/paper/2023/hash/523f87e9d08e6071a3bbd150e6da40fb-Abstract-mlsys2023.html} {Efficiently scaling transformer inference}.
\newblock \emph{Proceedings of Machine Learning and Systems}, 5.

\bibitem[{Post(2018)}]{sbleu}
Matt Post. 2018.
\newblock \href {https://doi.org/10.18653/v1/W18-6319} {A call for clarity in reporting {BLEU} scores}.
\newblock In \emph{Proceedings of the Third Conference on Machine Translation: Research Papers}, pages 186--191, Brussels, Belgium. Association for Computational Linguistics.

\bibitem[{Productions(2023)}]{slang}
Sharpened Productions. 2023.
\newblock Slang.net: The slang dictionary.
\newblock \url{https://slang.net/}.
\newblock Accessed: 2023-08-14.

\bibitem[{Qiu et~al.(2020)Qiu, Sun, Xu, Shao, Dai, and Huang}]{qiu2020pre}
Xipeng Qiu, Tianxiang Sun, Yige Xu, Yunfan Shao, Ning Dai, and Xuanjing Huang. 2020.
\newblock \href {https://doi.org/https://doi.org/10.1007/s11431-020-1647-3} {Pre-trained models for natural language processing: A survey}.
\newblock \emph{Science China Technological Sciences}, 63(10):1872--1897.

\bibitem[{Radford et~al.(2019)Radford, Wu, Child, Luan, Amodei, and Sutskever}]{radford2019language}
Alec Radford, Jeff Wu, Rewon Child, David Luan, Dario Amodei, and Ilya Sutskever. 2019.
\newblock Language models are unsupervised multitask learners.

\bibitem[{Raffel et~al.(2020)Raffel, Shazeer, Roberts, Lee, Narang, Matena, Zhou, Li, and Liu}]{raffel2020exploring}
Colin Raffel, Noam Shazeer, Adam Roberts, Katherine Lee, Sharan Narang, Michael Matena, Yanqi Zhou, Wei Li, and Peter~J. Liu. 2020.
\newblock \href {http://jmlr.org/papers/v21/20-074.html} {Exploring the limits of transfer learning with a unified text-to-text transformer}.
\newblock \emph{Journal of Machine Learning Research}, 21(140):1--67.

\bibitem[{Rao and Tetreault(2018)}]{gyafc}
Sudha Rao and Joel Tetreault. 2018.
\newblock \href {https://doi.org/10.18653/v1/N18-1012} {Dear sir or madam, may {I} introduce the {GYAFC} dataset: Corpus, benchmarks and metrics for formality style transfer}.
\newblock In \emph{Proceedings of the 2018 Conference of the North {A}merican Chapter of the Association for Computational Linguistics: Human Language Technologies, Volume 1 (Long Papers)}, pages 129--140, New Orleans, Louisiana. Association for Computational Linguistics.

\bibitem[{Rei et~al.(2020)Rei, Stewart, Farinha, and Lavie}]{comet}
Ricardo Rei, Craig Stewart, Ana~C Farinha, and Alon Lavie. 2020.
\newblock \href {https://doi.org/10.18653/v1/2020.emnlp-main.213} {{COMET}: A neural framework for {MT} evaluation}.
\newblock In \emph{Proceedings of the 2020 Conference on Empirical Methods in Natural Language Processing (EMNLP)}, pages 2685--2702, Online. Association for Computational Linguistics.

\bibitem[{Touvron et~al.(2023)Touvron, Lavril, Izacard, Martinet, Lachaux, Lacroix, Rozière, Goyal, Hambro, Azhar, Rodriguez, Joulin, Grave, and Lample}]{touvron2023llama}
Hugo Touvron, Thibaut Lavril, Gautier Izacard, Xavier Martinet, Marie-Anne Lachaux, Timothée Lacroix, Baptiste Rozière, Naman Goyal, Eric Hambro, Faisal Azhar, Aurelien Rodriguez, Armand Joulin, Edouard Grave, and Guillaume Lample. 2023.
\newblock \href {http://arxiv.org/abs/2302.13971} {Llama: Open and efficient foundation language models}.

\bibitem[{{\"U}st{\"u}n and Cooper~Stickland(2022)}]{ustun-2022-parameter}
Ahmet {\"U}st{\"u}n and Asa Cooper~Stickland. 2022.
\newblock \href {https://doi.org/10.18653/v1/2022.emnlp-main.540} {When does parameter-efficient transfer learning work for machine translation?}
\newblock In \emph{Proceedings of the 2022 Conference on Empirical Methods in Natural Language Processing}, pages 7919--7933, Abu Dhabi, United Arab Emirates. Association for Computational Linguistics.

\bibitem[{Xiao et~al.(2023)Xiao, Lin, and Han}]{xiao2023offsite}
Guangxuan Xiao, Ji~Lin, and Song Han. 2023.
\newblock \href {http://arxiv.org/abs/2302.04870} {Offsite-tuning: Transfer learning without full model}.

\bibitem[{Yahoo(2007)}]{yahoo}
Yahoo. 2007.
\newblock L6 - yahoo! answers comprehensive questions and answers version 1.0.
\newblock \url{https://webscope.sandbox.yahoo.com/}.
\newblock Accessed: 2023-07-02.

\bibitem[{Zhang et~al.(2023)Zhang, Chen, Bukharin, He, Cheng, Chen, and Zhao}]{zhang2023adaptive}
Qingru Zhang, Minshuo Chen, Alexander Bukharin, Pengcheng He, Yu~Cheng, Weizhu Chen, and Tuo Zhao. 2023.
\newblock \href {https://openreview.net/forum?id=lq62uWRJjiY} {Adaptive budget allocation for parameter-efficient fine-tuning}.
\newblock In \emph{The Eleventh International Conference on Learning Representations}.

\bibitem[{Zhang et~al.(2022)Zhang, Roller, Goyal, Artetxe, Chen, Chen, Dewan, Diab, Li, Lin, Mihaylov, Ott, Shleifer, Shuster, Simig, Koura, Sridhar, Wang, and Zettlemoyer}]{zhang2022opt}
Susan Zhang, Stephen Roller, Naman Goyal, Mikel Artetxe, Moya Chen, Shuohui Chen, Christopher Dewan, Mona Diab, Xian Li, Xi~Victoria Lin, Todor Mihaylov, Myle Ott, Sam Shleifer, Kurt Shuster, Daniel Simig, Punit~Singh Koura, Anjali Sridhar, Tianlu Wang, and Luke Zettlemoyer. 2022.
\newblock \href {http://arxiv.org/abs/2205.01068} {Opt: Open pre-trained transformer language models}.

\end{thebibliography}
\appendix

\section{Performance on Different Rank Sizes}
\begin{table}[h]
    \centering
    \resizebox{1\linewidth}{!}{
        \begin{tabular}{lrrr}
            \toprule
            Model & \multicolumn{1}{c}{WMT22} & \multicolumn{1}{c}{GYAFC} & \multicolumn{1}{c}{GYAFC}\\
 & \multicolumn{1}{c}{(EN$\rightarrow$DE)} & \multicolumn{1}{c}{(F\&R)} & \multicolumn{1}{c}{(E\&M)}\\
\midrule
OPT-1.3B (LoRA${}_{r=8}$) & 3.25&23.13&18.41\\
OPT-1.3B (LoRA${}_{r=512}$) & 4.28&20.98&15.57\\
OPT-1.3B (PEMA${}_{r=8}$) & {\underline{11.75}}&\underline{56.29}&\underline{54.22}\\
OPT-1.3B (PEMA${}_{r=512}$) & \textbf{12.87}&\textbf{64.82}&\textbf{61.24}\\
\midrule
LLaMA-7B (LoRA${}_{r=8}$) & 10.92&14.80&12.69\\
LLaMA-7B (LoRA${}_{r=512}$) & {\underline{11.46}}&\underline{52.67}&\underline{52.15}\\
LLaMA-7B (PEMA${}_{r=8}$) & 3.88&48.88&45.73\\
LLaMA-7B (PEMA${}_{r=512}$) & \textbf{14.50}&\textbf{63.99}&\textbf{60.88}\\
\midrule
OPT-30B (LoRA${}_{r=8}$)& 16.05&61.28&59.48\\
OPT-30B (LoRA${}_{r=512}$)& 16.03&61.39&59.76\\
OPT-30B (PEMA${}_{r=8}$)& {\underline{18.33}}&\underline{62.87}&\underline{60.12}\\
OPT-30B (PEMA${}_{r=512}$)& \textbf{19.22}&\textbf{70.84}&\textbf{65.43}\\
\bottomrule
        \end{tabular}
    }
    \caption{Experiment on LoRA and \sys on meaning preservation (sBLEU) across rank variations ($r=\{8, 512\}$). The result shows \sys consistently outperforms LoRA on sBLEU and COMET.}
    \label{tab:model_comparison_rank}
\end{table}

LoRA~\cite{hu2022lora} states performance remains comparable with a small rank size.
However, AdaLoRA~\cite{zhang2023adaptive} finds a large rank size in the last layer of PLMs is needed for better performance.
Performance evaluation on \sys and baseline PEFT methods is conducted at the last layer of PLMs.
For this reason, we set $r=512$ for LoRA and \sys to minimize the effect on performance with rank size.
However, LoRA uses a rank size between 1 to 64 for their experiment.
As \sys is a LoRA-based PEFT method, we compared the performance on meaning preservation using the rank size employed in LoRA (8) and the rank size used in our experiment (512).
As Table~\ref{tab:model_comparison_rank} shows, a larger rank size generally achieves favorable performance. 
In the case of LoRA, using a rank size of 512 outperforms 8 in 6 out of 9 cases.
\sys with a rank size of 512 performs better than \sys with a rank size of 8 at all tasks.

\section{Measuring Informal Language Patterns}

\begin{table*}[t!]
    \centering
    \resizebox{0.9\linewidth}{!}{
        \begin{tabular}{lrrrrrrrr}
            \toprule
                          & \multicolumn{1}{c}{Informal} & \multicolumn{1}{c}{Formal} & \multicolumn{1}{c}{Naïve} & \multicolumn{1}{c}{$k$NN-LM} & \multicolumn{1}{c}{LoRA} & \multicolumn{1}{c}{UniPELT} & \multicolumn{1}{c}{Offsite-Tuning} & \multicolumn{1}{c}{PEMA} \\
                        & \multicolumn{1}{c}{Input}    & \multicolumn{1}{c}{Reference} & \multicolumn{1}{c}{OPT-30B}              & \multicolumn{1}{c}{}       & \multicolumn{1}{c}{}     & \multicolumn{1}{c}{}      & \multicolumn{1}{c}{}   & \multicolumn{1}{c}{}     \\
\midrule
Family \& Relationships &                              &                            &                                   &                            &                          &                             &                          \\
\midrule
Slang Abbreviation      & 525                          & 307.75                     & 346                               & 339                        & 356                      & 322                         & 361 & \textbf{289}                      \\
All Capital             & 68                           & 0                          & 61                                & 60                         & 8                        & 5                           & 65 & \textbf{3}                        \\
Redundant Word          & 39                           & 2                          & 1                                 & 1                          & 2                        & \textbf{0}                  & 17 & 3                        \\
Non-Capital Start       & 636                          & 1.5                        & 16                                & 2                          & 1                        & 1                           & 2 & \textbf{0}                        \\
\midrule
Entertainment \& Music  &                              &                            &                                   &                            &                          &                             & &                          \\
\midrule
Slang Abbreviation      & 651                          & 485.75                     & 541                               & 538                        & 530                      & 534                         & 529 & \textbf{463}                      \\
All Capital             & 36                           & 0                          & 31                                & 34                         & 9                        & 9                           & 37 & \textbf{0}                        \\
Redundant Word          & 49                           & 17.75                      & 5                                 & 5                          & 7                        & \textbf{3}                  & 16 & 32                       \\
Non-Capital Start       & 655                          & 7                          & 24                                & 2                          & \textbf{0}                        & 1                  & 3 & \textbf{0}                       \\
\bottomrule
        \end{tabular}
    }
    \caption{Count of informal patterns for each generated formal sentence. The result shows that \sys performs better in mitigating informal patterns than baseline approaches. Lower is better.}
    \label{tab:pattern_ex}
\end{table*}

The GYAFC dataset for style transfer includes common informal input patterns that are frequently occur.
To analyze the amount of mitigation, we categorize these patterns into four types.
The four informal patterns are as follows.
\textbf{Slang abbreviations} are informal short forms of words or phrases (e.g., "LOL"-"laughing out loud"). 
To identify the presence of slang words, we check how many words from the predicted target sentence are present in the slang dictionary from~\cite{slang}.
\textbf{All capital} is a pattern in which all characters in a generated word are capitalized (e.g., "FUNNY"). 
We calculate how many generated words are all capitalized.
\textbf{Redundant word} occurs when two consecutive words are the same. For example, "I lie lie lie and then I lie some more." has two redundant words.
\textbf{Non-capital start} is counted when a sentence does not start with a capital letter (e.g., "i only want points"). 

Table~\ref{tab:pattern_ex} shows the count of each informal pattern in generated sentences for both the baseline and \sys.
We also show an informal pattern count on informal input and formal reference.
There are four reference sentences for each example in the test set. 
We show the average count for each pattern using the formal reference.
It shows \sys is good at mitigating slang abbreviation, all capital, and non-capital start compared to other baseline approaches.
Interestingly, \sys outperforms formal references in mitigating slang abbreviations and non-capital start.

\section{Dataset}
\subsection{Data Statistic}
Table~\ref{tab:data_stat} shows data statistics of GYAFC and WMT22. 
For WMT22, we use a news-commentary v16 (EN$\rightarrow$DE) for training. The test set for GYAFC has four references, while WMT22 has one reference for each test input.

\begin{table}[t]
    \centering
    \resizebox{0.99\linewidth}{!}{
        \begin{tabular}{lrrrr}
            \toprule
            Dataset                        & \multicolumn{1}{c}{Train}     & \multicolumn{1}{c}{Valid}            & \multicolumn{1}{c}{Test} & \multicolumn{1}{c}{Length of $\mathcal{E}$}\\
\midrule
GYAFC (F\&R) & \multicolumn{1}{r}{51,967} &\multicolumn{1}{r}{2,788}& \multicolumn{1}{r}{1,332}       & \multicolumn{1}{r}{691,531}     \\
GYAFC (E\&M)  & \multicolumn{1}{r}{52,595} &\multicolumn{1}{r}{2,877}& \multicolumn{1}{r}{1,416}       & \multicolumn{1}{r}{695,465}      \\
WMT22  & \multicolumn{1}{r}{388,482} & \multicolumn{1}{r}{2,203} & \multicolumn{1}{r}{1,984}       & \multicolumn{1}{r}{20,983,482}     \\
\bottomrule
        \end{tabular}
    }
    \caption{Data statistic of GYAFC and WMT22 with length of external memory $\mathcal{E}$.}
    \label{tab:data_stat}
\end{table}
\subsection{Dataset Examples}
Table~\ref{tab:dataset_example} demonstrates examples of parallel datasets of GYAFC and WMT22.
\begin{table}[t]
    \centering
    \resizebox{0.99\linewidth}{!}{
        \begin{tabular}{lll}
            \toprule
            Task  & Example &\\
\midrule
WMT22 & English: &In better shape, but not alone.\\
      & German: &In besserer Verfassung, aber nicht allein.\\
\midrule
GYAFC & Informal: &I’d say it is punk though.\\
      & Formal: &However, I do believe it to be punk.\\
\bottomrule
        \end{tabular}
    }
    \caption{Example of parallel dataset GYAFC and WMT22.}
    \label{tab:dataset_example}
\end{table}

\subsection{Prompts}
Table~\ref{tab:prompt} presents prompt input used for evaluation. 
WMT22 and GYAFC have two placeholders.
This includes [English Input] and [Informal Input].
[Generated Output] is a predicted output sentence generated by PLMs.

[English Input] represents the English input sentence in WMT22. 
[Informal Input] is the informal input sentence in GYAFC.
An example of the parallel data input can be found in Table~\ref{tab:dataset_example}.
 
\begin{table}[t!]
    \centering
    \resizebox{0.99\linewidth}{!}{
        \begin{tabular}{ll}
            \toprule
            Task  & Prompt                                                     \\
\midrule
WMT22 & Translate this from English to German:                          \\
      & [English Input]                                                 \\
      & German: [Generated Output]                                      \\
\midrule
GYAFC & Convert the following informal sentence into a formal sentence: \\
      & Informal: [Informal Input]                                      \\
      & Formal: [Generated Output]                                      \\
\bottomrule
        \end{tabular}
    }
    \caption{Prompt used for evaluation. [ ] represents the placeholder.}
    \label{tab:prompt}
\end{table}

\subsection{Post-processing}
We use three decoder-based pre-trained language models for evaluation: OPT-IML-MAX-1.3B, LLaMA-7B, and OPT-IML-MAX-30B.
These models are capable of generating tokens continuously.
This characteristic makes decoder-based language models generate beyond the predicted sentences, typically called hallucinations.
We find common hallucination patterns in each pre-trained language model.
We post-process hallucinations generated after the predicted sentence for evaluation.
Table~\ref{tab:common_hallucinations} shows common hallucination patterns that are removed.

\begin{table}[t!]
    \centering
    \resizebox{0.9\linewidth}{!}{
        \begin{tabular}{ll}
            \toprule
            Model & Common hallucination patterns \\
\midrule
OPT & I'm not sure … \\
 & I 50\% ... \\
 & Convert the following informal   sentence … \\
 & Translate this from English to   German: … \\
 & I. … \\
 & ….. … \\
\midrule
LLaMA & Informal: … \\
 & \#\#\# … \\
 & Comment: … \\
 & \textbackslash{}\textbackslash{}[ … \\
 & \textbackslash{}begin ... \\
 & Answer: … \\
\bottomrule

        \end{tabular}
    }
    \caption{Common hallucination patterns after generating a predicted sentence.}
    \label{tab:common_hallucinations}
\end{table}

\section{Implementation Details}
We use three RTX 8000 GPUs with 48GB GDDR6 memory for our experiment.
For OPT-IML-MAX-1.3B, we use full precision (FP32) for training and inference.
For LLaMA-7B and OPT-IML-MAX-30B, we use half-precision (FP16) and distribute the model across three GPUs using the HuggingFace Accelerate library.
The hyperparameters for \sys and the baselines are in Table~\ref{tab:hyperparameter}. The best hyperparameter is selected using a grid search.
\begin{table}[t!]
    \centering
    \resizebox{0.8\linewidth}{!}{
        \begin{tabular}{lr}
            \toprule
            \sys& \multicolumn{1}{c}{} \\
\midrule
Random seed & 123\\
Batch size & 40,960 \\
Adam $lr$ & 1e-03 \\
Adam $(\beta_1, \beta_2)$& (0.9, 0.999) \\
Adam $eps$& 1e-08 \\
Number of rank& 512\\
Optimal $\lambda_{max}$ & 0.7 to 0.9\\
\midrule
\multicolumn{1}{l}{Offsite-Tuning}& \multicolumn{1}{c}{} \\
\midrule
Random seed & 42 \\
Batch size & 18 \\
Emulator size & $\frac{1}{3}$ of PLM \\
Adam $lr$ & 1e-04 \\
Adam $(\beta_1, \beta_2)$& (0.9, 0.999) \\
Adam $eps$& 1e-08 \\
\midrule
\multicolumn{1}{l}{LoRA}& \multicolumn{1}{c}{} \\
\midrule
Random seed & 123\\
Batch size& 10 to 30\\
Adam $lr$ & 1e-03 \\
Adam $(\beta_1, \beta_2)$& (0.9, 0.999) \\
Adam $eps$& 1e-08 \\
Number of rank & 512\\
LoRA $\alpha$ & 1\\
Merge weight& FALSE\\
\midrule
\multicolumn{1}{l}{$k$NN-LM}& \multicolumn{1}{c}{} \\
\midrule
Random seed & 1\\
Number of centroids learn & 4,096 \\
Quantized vector size & 64 \\
Number of clusters to query & 32 \\
Distance function & L2 Distance\\
\midrule
\multicolumn{1}{l}{UniPELT} & \\
\midrule
Random seed & 123 \\
Batch size& 10 to 30\\
Adam $lr$ & 1e-03 \\
Adam $(\beta_1, \beta_2)$& (0.9, 0.999) \\
Adam $eps$& 1e-08 \\
Prefix gate& True \\
Prefix length& 10 \\
Prefix mid dimension& 512 \\
LoRA gate & True \\
Number of rank & 10 \\
LoRA $\alpha$ & 16\\
Adapter gate & True \\
Adapter down sample & $D_{hid}/2$ \\
\multirow{3}{*}{Used PEFT methods} & Adapter \\
&Prefix tuning \\
&LoRA\\
\bottomrule
        \end{tabular}
    }
    \caption{Hyper-parameter setup of each baseline method. We select the batch size between 10 to 30. $D_{hid}$ represent hidden size of a model.}
    \label{tab:hyperparameter}
\end{table}

\section{Examples of Generated Outputs}
The generated formal outputs of GYAFC are shown in Table~\ref{tab:gyafc_em_examples} and Table~\ref{tab:gyafc_fr_examples}.
In WMT22, the German output generated is presented in Table~\ref{tab:wmt22_examples}. It shows \sys understands the meaning of abbreviated format (e.g., translating "5'4" to "5 feet 4 inches"), or removing the informal word (e.g., "flirt" which typically refers to playful or teasing behavior). Mitigating common informal patterns such as all capital words (e.g., "PINK FLOYD" to "Pink Floyd") while preserving the meaning of input (e.g., "Wir" means "We" in German). 

\section{Difference Between \sys and LoRA at $W_{hd}$} 
Applying LoRA to $W_{hd} \in \mathbb{R}^{v \times d}$, a larger set of parameters is required due to the difference in input and output sizes ($d$ and $v$). Conversely, \sys operates more efficiently, utilizing computation resources by receiving an input of size $d$ and yielding an output of the same size. For instance, OPT-1.3B has $d=2,048$ and $v=50,272$.

\begin{figure}[!t]
    \centering
    \includegraphics[width=0.9\linewidth]{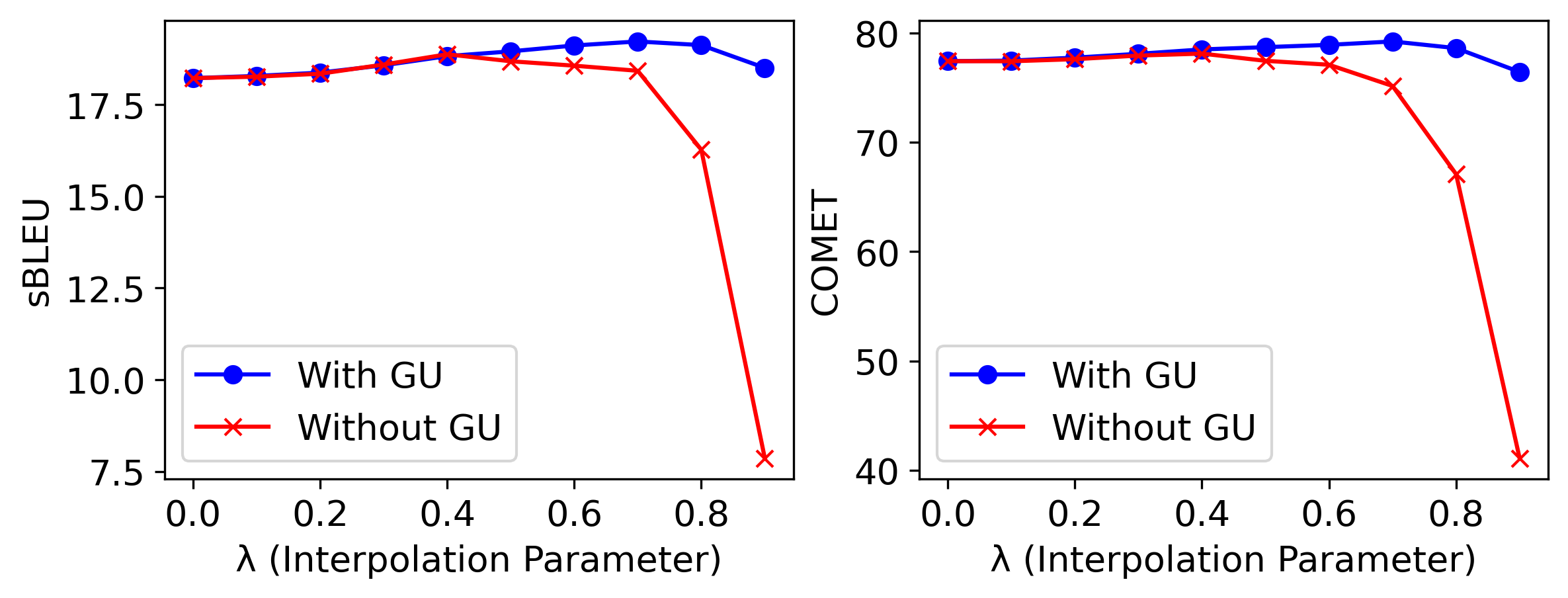}
    \caption{
    Performance variation for each interpolation value $\lambda_{max}$ in the WMT22 task. 
    With both Gradual Unrolling ($GU$) (blue) and without $GU$ (red), there is a decline in performance at a specific point of $\lambda_{max}$.
    However, when utilizing $GU$, the model is not only robust to performance degradation but also gains performance improvement.}
    \label{fig:lambda}
\end{figure}

\begin{figure}[!t]
    \centering
    \includegraphics[width=0.9\linewidth]{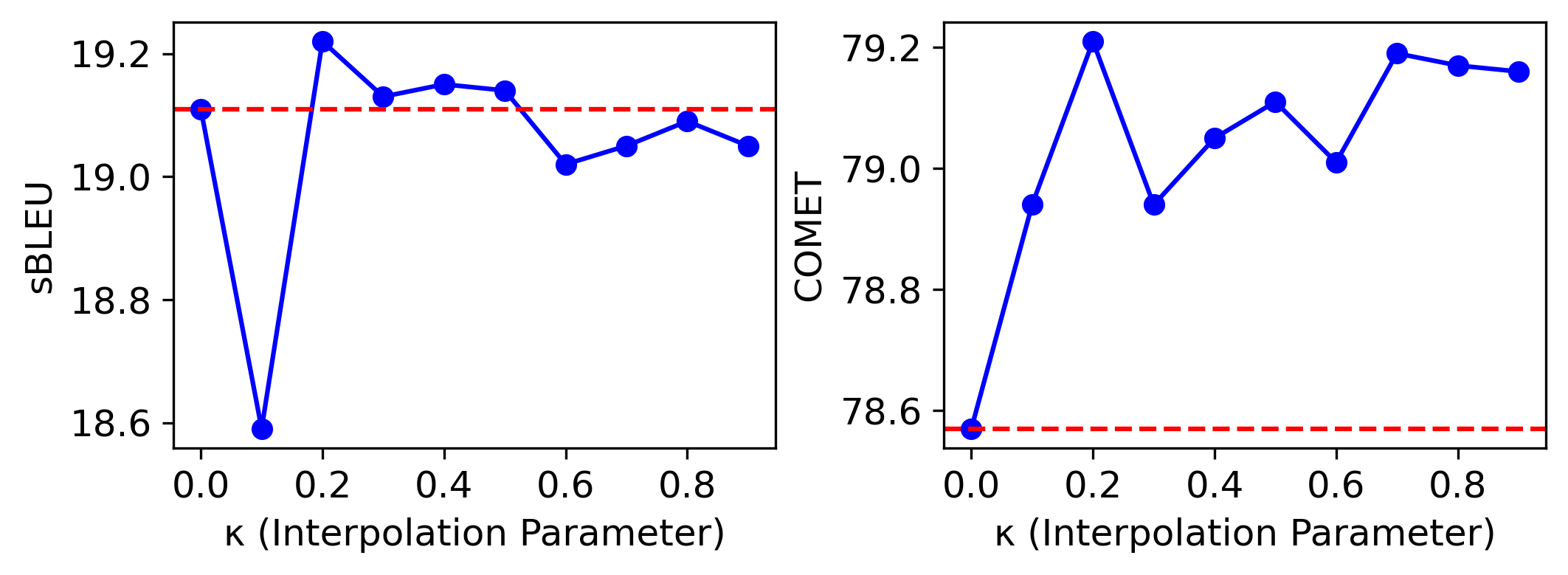}
    \caption{
    Impact of mixing ratio values between reconstruction loss and predicting the next-token loss in the WMT22 task.
    When $\kappa$ is $0$, it means excluding reconstruction loss (red dashed line).
    We fix the $\lambda_{max}$ value as 0.7.
    The graphs show that combining reconstruction loss and predicting the next-token loss is superior to excluding reconstruction loss.}
    \label{fig:kappa_wmt}
\end{figure}

\section{Impact on Interpolation $\lambda$ and $\kappa$}
In the WMT22 task, we observe performance variation with different interpolation values, $\lambda_{max}$ in Figure~\ref{fig:lambda}. Additionally, we investigate the impact of the mixing ratio values between reconstruction loss and predicting the next-token loss in Figure~\ref{fig:kappa_wmt}.

\section{Rule of Thumb to Choose $\kappa$}
The training process of \sys consists of two distinct phases. Initial reconstruction training and joint retraining. Because they both have user-defined variables $\kappa$ and $\lambda$, It may be hard to tune both variables to find optimal performance. Especially, because $\kappa$ is defined at initial reconstruction training, it may be difficult to train separate models for different $\kappa$.
Thus, we show the rule of thumb of choosing the $\kappa$. 

In our experiments with tasks such as WMT22 and GYAFC, we found that $\kappa$ values between 0.2 and 0.5 yielded the best results.
Figure~\ref{fig:kappa_wmt} clearly shows that \sys reached the optimal $\kappa$ value quite early, around 0.2, and observed a noticeable decline in performance, particularly when the value exceeded 0.5. We also reveal the impact of the interpolation value $\kappa$ on the GYAFC task, as presented in Table~\ref{tab:optimal_interpolation_kappa}, which aligns with our findings. Based on these observations, we propose a rule of thumb for selecting $\kappa$ should consider a range between 0.2 and 0.5. This range balances performance and efficiency well across the tasks we evaluated.

\begin{table*}[t!]
    \centering
    \resizebox{0.9\linewidth}{!}{
        \begin{tabular}{llllllllllll}
            \toprule
            \textbf{}  & {Interpolation ($\kappa$)} & \multicolumn{1}{c}{{0.0}} & \multicolumn{1}{c}{{0.1}} & \multicolumn{1}{c}{{0.2}}      & \multicolumn{1}{c}{{0.3}}      & \multicolumn{1}{c}{{0.4}} & \multicolumn{1}{c}{{0.5}}      & \multicolumn{1}{c}{{0.6}} & \multicolumn{1}{c}{{0.7}}      & \multicolumn{1}{c}{{0.8}} & \multicolumn{1}{c}{{0.9}} \\
\midrule
GYAFC (EM) & sBLEU                      & 65.21                            & 64.52                            & 64.69                                 & \textbf{65.43} & 64.22                            & 65.13                                 & 64.53                            & 64.98                                 & 65.19                            & 65.03                            \\
           & FormImp                    & 44.05                            & 42.12                            & 44.93                                 & 44.63                                 & 43.09                            & \textbf{45.15} & 44.13                            & 44.04                                 & 44.06                            & 44.40                            \\
\midrule
GYAFC (FR) & sBLEU                      & 70.40                            & 70.42                            & \textbf{70.84} & 70.78                                 & 70.36                            & 70.08                                 & 70.20                            & 70.63                                 & 70.76                            & 70.76                            \\
           & FormImp                    & 52.50                            & 51.40                            & 52.35                                 & 52.05                                 & 51.79                            & 51.37                                 & 52.46                            & \textbf{52.78} & 51.20                            & 51.83\\
\bottomrule
        \end{tabular}
    }
    \caption{Impact of interpolation value $\kappa$ on GYAFC with OPT-IML-MAX-30B. Our finding shows the optimal $\kappa$ is mostly within a range between 0.2 and 0.5.}
    \label{tab:optimal_interpolation_kappa}
\end{table*}

\section{Evaluating the Impact of Gradual Unrolling on $k$NN-LM}
The Gradual Unrolling strategy is applicable across baselines that interpolate between two distributions of the next-token. This means the GU can be applied to the $k$NN-LM baseline. We conducted a comparative analysis in Table~\ref{tab:comparison_with_gu} to demonstrate the effectiveness of GU by comparing the performance of $k$NN-LM and \sys with and without the GU. The result shows that \sys consistently outperforms the $k$NN-LM approach, even when the GU is applied to the $k$NN-LM.

\begin{table*}[t!]
    \centering
    \resizebox{0.9\linewidth}{!}{
        \begin{tabular}{lrrrrrrrrr}
            \toprule
            \textbf{}                 & \multicolumn{3}{c}{{WMT22 (EN$\rightarrow$DE)}}                          & \multicolumn{3}{c}{{GYAFC (F\&R)}}                                         & \multicolumn{3}{c}{{GYAFC (E\&M)}}                                         \\
                          & \multicolumn{1}{c}{sBLEU} & \multicolumn{1}{c}{PPL} & \multicolumn{1}{c}{COMET} & \multicolumn{1}{c}{sBLEU} & \multicolumn{1}{c}{PPL} & \multicolumn{1}{c}{FormImp} & \multicolumn{1}{c}{sBLEU} & \multicolumn{1}{c}{PPL} & \multicolumn{1}{c}{FormImp} \\
\midrule
OPT-1.3B ($k$NN-LM)          & 8.07                      & 91.37                   & 41.75                     & \underline{56.69}               & \underline{20.87}             & \underline{16.26}                 & 54.74                     & 23.15                   & 14.46                       \\
OPT-1.3B ($k$NN-LM with GU) & \underline{10.09}               & \underline{51.82}             & \underline{56.57}               & 56.21                     & \textbf{19.68}          & 9.73                        & \underline{55.21}               & \textbf{19.69}          & \underline{23.43}                 \\
OPT-1.3B (PEMA w/o GU)    & 9.39                      & 52.19                   & 56.36                     & 55.18                     & \textbf{19.68}          & 9.38                        & 53.73                     & \underline{21.47}             & 8.62                        \\
OPT-1.3B (PEMA)           & \textbf{12.87}            & \textbf{42.62}          & \textbf{64.16}            & \textbf{64.82}            & 23.15                   & \textbf{41.90}              & \textbf{61.24}            & 24.28                   & \textbf{36.28}             \\
\bottomrule

        \end{tabular}
    }
    \caption{Comparative analysis of \sys and $k$NN-LM with and without GU implementation. The default configuration of PEMA incorporates GU. Hence, we report PEMA except for GU as 'PEMA w/o GU.'}
    \label{tab:comparison_with_gu}
\end{table*}

\section{Evaluation Beyond Zero-shot Inference}
We conducted all experiments based on zero-shot inference. However, zero-shot inference might not show the robustness of the results when few-shot in-context learning is applied. To validate its robustness, we conducted an experiment with few-shot in-context learning. We used LLaMA 7B as a baseline and provided five-shot examples at inference. We compared naïve LLaMA 7B and LLaMA 7B with LoRA as baselines and compared baselines with \sys. The result is shown in Table~\ref{tab:few_shot_result_llama}. The result shows that few-shot in-context learning benefits performance in sBLEU across all methods.

\section{Investigation Given Paraphrased Input}
One interesting aspect of \sys is that it allows the data owner to determine the amount of data provided to the PLM owner for initial input. 
For example, in a parallel dataset, the initial input might differ from the source input in the original data (i.e., which the data owner holds) but convey a similar meaning. 
To understand its performance given paraphrased inputs, we use a Mixtral-8x7B-Instruct~\cite{jiang2024mixtral} to paraphrase the informal sentences from the initial prompt in the GYAFC dataset. 
Table~\ref{tab:paraphrased} shows examples of paraphrases generated by Mixtral-8x7B-Instruct. The examples include well-paraphrased and challenging examples, all of which we used for evaluation. Afterward, we use the prompt from Table~\ref{tab:prompt} and only switch [Informal Input] to [Paraphrased Informal Input].

This ensures that the paraphrased initial input, rather than the original input, is provided to the PLM. We then input this data into OPT-IML-MAX-1.3B to gather context representation. Subsequently, we construct an external memory to train \sys. The test set remains unchanged for an accurate performance comparison. Table~\ref{tab:performance_paraphrased} shows the performance between the paraphrased and original inputs. Note that only "\sys with PI" used the paraphrased input, while the others used the original data for training. The results show that the performance of \sys with paraphrased input is slightly lower than that with the original data (about 4 to 5 sBLEU). Interestingly, \sys with PI still surpasses baselines that utilize original input\footnote{Please refer to Table~\ref{tab:model_comparison_1} to compare with other baselines.}.

\section{Licensing Information}
\noindent
\textbf{Models}
OPT is licensed under the MIT License. The LLaMA is licensed under the GNU General Public License (GPL) version 3.\\
\noindent
\textbf{Fine-tuning Methods}
$k$NN-LM, LoRA, and Offsite-Tuning are licensed under the MIT License. UniPELT is licensed under the Creative Commons Attribution-NonCommercial (CC-BY-NC) license.\\
\noindent
\textbf{Dataset}
GYAFC is based on the Yahoo Answers corpus (L6 - Yahoo! Answers Comprehensive Questions and Answers version 1.0)~\cite{yahoo}, and is designated for research purposes. Access to the GYAFC dataset requires access to Yahoo Answers corpus.
WMT22 is freely available for academic and educational research.

\begin{table*}[t]
    \centering
    \resizebox{1\linewidth}{!}{
        \begin{tabular}{lrrrrrrrrr}
            \toprule
            \textbf{}       & \multicolumn{3}{c}{{WMT22 (EN$\rightarrow$DE)}}                          & \multicolumn{3}{c}{{GYAFC (F\&R)}}                                         & \multicolumn{3}{c}{{GYAFC (E\&M)}}                                         \\
                & \multicolumn{1}{c}{sBLEU} & \multicolumn{1}{c}{PPL} & \multicolumn{1}{c}{COMET} & \multicolumn{1}{c}{sBLEU} & \multicolumn{1}{c}{PPL} & \multicolumn{1}{c}{FormImp} & \multicolumn{1}{c}{sBLEU} & \multicolumn{1}{c}{PPL} & \multicolumn{1}{c}{FormImp} \\
\midrule
LLaMA 7B        & 7.84                      & 41.42                   & 59.47                     & 24.50                     & 36.62                   & 52.63                       & 27.41                     & 39.15                   & 66.12                       \\
LLaMA 7B (LoRA) & 17.60                     & 38.58                   & \textbf{78.71}            & 55.07                     & \textbf{22.77}          & 18.18                       & 51.02                     & \textbf{25.93}          & 19.26                       \\
LLaMA 7B (PEMA) & \textbf{17.75}            & \textbf{37.27}          & 77.01                     & \textbf{65.01}            & 24.71                   & \textbf{63.47}              & \textbf{65.68}            & 27.23                   & \textbf{76.91}             \\
\bottomrule
        \end{tabular}
    }
    \caption{Comparison of different tasks on few-shot in-context learning using LLaMA-7B. All results are from LLaMA 7B with five-shot examples.}
    \label{tab:few_shot_result_llama}
\end{table*}

\begin{table*}
    \centering
    \resizebox{1\linewidth}{!}{
        \begin{tabular}{ll}
            \toprule
            {Original Informal}   & {Paraphrased by Mixtral-8x7B-Instruct}              \\
\midrule
IT WAS SAD AT THE END.       & THE END WAS DISMAL.                               \\
Er have you heard of google? & hrsLECBECLECBECLECBECBECBECBECBECBECBECBECBECBE  … \\
he was called sleepy k n o b & he was referred to as drowsy k n o b              \\
fall out boy b/c they rock   & of fun w/ fall out boy b/c they're awesome      \\
\bottomrule
        \end{tabular}
    }
    \caption{Examples of original input and paraphrased by Mixtral-8x7B-Instruct on the GYAFC dataset.}
    \label{tab:paraphrased}
\end{table*}

\begin{table*}
    \centering
    \resizebox{0.8\linewidth}{!}{
        \begin{tabular}{lrrrrrr}
            \toprule
            \textbf{}                             & \multicolumn{3}{c}{{GYAFC (F\&R)}}                                         & \multicolumn{3}{c}{{GYAFC (E\&M)}}                                         \\
                                      & \multicolumn{1}{c}{sBLEU} & \multicolumn{1}{c}{PPL} & \multicolumn{1}{c}{FormImp} & \multicolumn{1}{c}{sBLEU} & \multicolumn{1}{c}{PPL} & \multicolumn{1}{c}{FormImp} \\
\midrule
OPT-1.3B                              & 55.00          & \textbf{18.98} & 11.05          & 53.98          & \textbf{20.89} & 10.67          \\
OPT-1.3B (Offsite-Tuning)             & 59.01          & 20.70          & 24.82          & 57.01          & 23.25          & 23.76          \\
OPT-1.3B (PEMA)                       & \textbf{64.82} & 23.15          & \textbf{41.90} & \textbf{61.24} & 24.28          & \textbf{36.28} \\
OPT-1.3B (PEMA with PI) & 59.86          & 21.76          & 26.78          & 57.02          & 23.28          & 24.47   \\
\bottomrule
        \end{tabular}
    }
    \caption{Performance comparison of \sys and baselines with paraphrased and original input in GYAFC.}
    \label{tab:performance_paraphrased}
\end{table*}

\begin{table*}[t]
    \centering
    \resizebox{1.0\linewidth}{!}{
        \begin{tabular}{lllr}
            \toprule
            Input     &               & he is probably wondering if your interested in him at all....flirt back!!                         & sBLEU  \\
\midrule
Reference & 1             & He is likely wondering if you are interested in him at all; Flirt back   with him.                &        \\
          & 2             & He probably wants to know if you're interested in him.                                            &        \\
          & 3             & He is probably wondering if you are interested in him at all, so flirt   back.                    &        \\
          & 4             & He is probably wondering if you are interested in him at all. Flirt back.                         &        \\
\midrule
Output    & PEMA          & He is probably wondering if you are interested in him at all.                                     & 100.0  \\
          & LoRA          & He is probably wondering if you are interested in him at all.  If you are interested, flirt back. & 66.78  \\
          & $k$NN-LM        & It is most likely that he is wondering if you are interested in him at   all....flirt back!!      & 42.60  \\
          & UniPELT       & He is probably wondering if your interested in him at all....flirt   back!                        & 50.82  \\
          & Offsite-Tuning & He probably is wondering if you are interested in him at all.  Flirt back!! & 72.98\\
          & Naïve OPT-30B & In informal situations he is probably wondering if your interested in him   at all.               & 46.03  \\
\bottomrule
\toprule
Input     &               & I don't know!...I just want the points...lol                                     &        \\
\midrule
Reference & 1             & I only want points.         &        \\
          & 2             & I do not know. I merely want the points.         &        \\
          & 3             & I do not know; I just want the points.         &        \\
          & 4             & I do not know, I only want the points.         &        \\
\midrule
Output    & PEMA          & I do not know, but I just want the points.                        & 73.49\\
          & LoRA          & I don't know!... I just want the points.  I am not sure what I am doing.& 25.31\\
          & $k$NN-LM        & I don't know!...I just want the points...lol                        & 34.90\\
          & UniPELT       & I don't know!...I just want the points...lol                        & 34.90\\
          & Offsite-Tuning & - & 0.00\\
          & Naïve OPT-30B & I don't know!...I just want the points...lol                        & 34.90\\
\bottomrule
\toprule
Input     &               & No way im 5`4 and he`s 6`2                                                                        &        \\
\midrule
Reference & 1             & No, I am 5ft 4inches and he is 6ft and 2inches.                                                   &        \\
          & 2             & No way, I am only 5'4" and he is 6'2".                                                            &        \\
          & 3             & Not at all. I am five feet four inches tall and he is 6 feet 2 inches   tall. `                   &        \\
          & 4             & No chance, I am five feet four inches tall and he is six feet two inches   tall.                  &        \\
\midrule
Output    & PEMA          & No way, I am 5 feet 4 inches tall and he is 6 feet 2 inches tall.                                 & 74.44  \\
          & LoRA          & No way, I am 5'4 and he is 6'2.                                                                   & 51.52  \\
          & $k$NN-LM        & No way, I am 5'4 and he is 6'2                                                                    & 50.05  \\
          & UniPELT       & No way, I am 5'4 and he is 6'2                                                                    & 50.05  \\
          & Offsite-Tuning & No way im 5'4 and he's 6'2. & 7.78 \\
          & Naïve OPT-30B & No way, I am 5`4 and he is 6`2                                                                    & 45.72 \\
\bottomrule
        \end{tabular}
    }
    \caption{Examples of generated formal output of GYAFC (Family \& Relationships) for given informal input. One interesting example is \sys can understand the meaning of abbreviated height descriptions like "5'4" and "6'2". And rewrite them into more formal forms "5 feet 4 inches" and "6 feet 2 inches".}
    \label{tab:gyafc_fr_examples}
\end{table*}
\begin{table*}[t]
    \centering
    \resizebox{1.0\linewidth}{!}{
        \begin{tabular}{lllr}
            \toprule
            Input     &               & I respect that tyra banks is trying show hosting..but the big O {[}aka   OPRAH{]} won...                         & \multicolumn{1}{c}{sBLEU} \\
\midrule
Reference & 1             & While I respect that Tyra Banks is branching out into television hosting,   Oprah Winfrey is the better example. &                           \\
          & 2             & I respect the hostess Tyra Banks, but I prefer Oprah.                                                            &                           \\
          & 3             & I admire Tyra Banks for trying to become a tv show host. However, Oprah   is clearly the best hostess.           &                           \\
          & 4             & I respect that Tyra Banks is trying out show hosting, but I think that   Oprah won.                              &                           \\
\midrule
Output    & PEMA          & I respect that Tyra Banks is trying to show hosting..but the big O {[}aka   Oprah Winfrey{]} won.                & 35.56                     \\
          & LoRA          & I respect that tyra banks is trying show hosting, but the big O won.                                             & 24.32                     \\
          & $k$NN-LM        & I respect that tyra banks is trying show hosting..but the big O {[}aka   OPRAH{]} won.                           & 12.02                     \\
          & UniPELT       & I respect that tyra banks is trying show hosting..but the big O   won...                                         & 13.40                     \\
          & Offsite-Tuning & I respect that tyra banks is trying show hosting..but the big O [aka OPRAH] won...& 10.90\\
          & Naïve OPT-30B & In informal: I respect that tyra banks is trying show hosting..but the   big O {[}aka OPRAH{]} won...            & 9.56                      \\
\bottomrule
\toprule
Input     &               & Not sure what's gonna happen, sure it will be good and OH YEAH!!!!!!!                                            &                           \\
\midrule
Reference & 1             & I am not sure what is going to happen but I am sure it will be good.                                             &                           \\
          & 2             & I'm not sure what is going to happen, I'm sure it will be good.                                                  &                           \\
          & 3             & I am not sure what will happen, but I am sure it will be good. Oh, yes.                                          &                           \\
          & 4             & I am not sure what will happen, however, I am sure that it will be good.                                         &                           \\
\midrule
Output    & PEMA          & I am not sure what is going to happen, but I am sure it will be good and   OH YEAH!!!!!!!                        & 61.11                     \\
          & LoRA          & Not sure what's going to happen, but I am sure it will be good and OH   YEAH!!!!!!!.                             & 43.05                     \\
          & $k$NN-LM        & Not sure what's gonna happen, sure it will be good and OH YEAH!!!!!!!                                            & 20.22                     \\
          & UniPELT       & Not sure what's going to happen, but                                                                             & 16.74                     \\
          & Offsite-Tuning & Not sure what's gonna happen, sure it will be good and OH YEAH!!!!!!! & 20.22 \\
          & Naïve OPT-30B & Not sure what's gonna happen, sure it will be good and OH YEAH!!!!!!!                                            & 20.22                     \\
\bottomrule
\toprule
Input     &               & Journey, PINK FLOYD, The POLICE, The EAGLES \& RUSH... omg!                                                      &                           \\
\midrule
Reference & 1             & Journey, Pink Floyd, The Police, The Eagles, and Rush - oh my!                                                   &                           \\
          & 2             & I like Journey, Pink Floyd, The Police, The Eagles, and Rush.                                                    &                           \\
          & 3             & Oh goodness, Journey, Pink Floyd, The Police, the Eagles, and Rush!                                              &                           \\
          & 4             & Journey, Pink Floyd, The Police, The Eagles, and Rush are all great   classic bands.                             &                           \\
\midrule
Output    & PEMA          & I love Journey, Pink Floyd, The Police, The Eagles and Rush.                                                     & 69.01                     \\
          & LoRA          & Journey, PINK FLOYD, The Police, The Eagles \& Rush.  I love it!                                                 & 36.45                     \\
          & $k$NN-LM        & Journey, PINK FLOYD, The Police, The Eagles \& Rush... omg!                                                      & 35.66                     \\
          & UniPELT       & Journey, PINK FLOYD, The Police, The Eagles \& Rush... omg!                                                      & 35.66                     \\
          & Offsite-Tuning & Journey, Pink Floyd, The Eagles, Rush, and The Police.  Oh my god!& 47.29 \\
          & Naïve OPT-30B & Journey, PINK FLOYD, The Police, The Eagles \& Rush... omg!                                                      & 35.66 \\
\bottomrule
        \end{tabular}
    }
    \caption{Examples of generated formal output of GYAFC (Entertainment \& Music) for given informal input. It shows that \sys is capable of restoring All Capital patterns to their formal format. For example, \sys successfully restore "Oprah Winfrey" given "OPRAH" as an input.}
    \label{tab:gyafc_em_examples}
\end{table*}
\begin{table*}[t]
    \centering
    \resizebox{1.0\linewidth}{!}{
        \begin{tabular}{lllr}
            \toprule
            Input     &               & In Berlin he trained as an opera singer and in 1986 he founded the Palast Orchester together with his classmates.&  \multicolumn{1}{c}{sBLEU}                         \\
\midrule
Reference &               & In Berlin studierte er Operngesang, 1986 gründete er zusammen mit Kommilitonen das Palast Orchester.         &                    \\
\midrule
Output    & PEMA          & In Berlin studierte er als Opernsänger und gründete 1986 mit seinen Schülern das Palast Orchester.           & 33.16                   \\
          & LoRA          & In Berlin wurde er als Opernsänger trainiert und 1986 gründete er mit seinen Schülern das Palast Orchester.  & 22.79                   \\
          & $k$NN-LM        & In Berlin schloss er 1986 mit seinen Schülern die Palast-Orchester zusammen.                                 & 6.64                   \\
          & UniPELT       & In Berlin trainierte er als Opernsänger und gründete 1986 mit seinen Schülern das Palast Orchester. When     & 19.35                   \\
          & Offsite-Tuning & ich war in Berlin als Opernstätte und 1986 gründete er die Palast Orchester gemeinsam mit seinen Kollegen.& 11.31\\
          & Naïve OPT-30B & In Berlin wurde er als Opernsänger trainiert und 1986 gründete er mit seinen Schülern das Palast Orchester.  & 22.79                   \\
\bottomrule
\toprule
Input     &               &  Since then the baritone has established himself with his repertoire of hits and chansons in the style of the Comedian Harmonists. &\\
\midrule
Reference &               &  Seitdem hat sich der Bariton mit seinem Repertoire aus Schlagern und Chansons im Stil der Comedian Harmonists etabliert. &\\
\midrule
Output    & PEMA          & Seitdem hat er sich mit seinem Repertoire aus Hits und Chansons im Stil der Comedian Harmonists etabliert.& 60.26                     \\
          & LoRA          & Seitdem ist er mit seinem Repertoire aus Hitparaden und Chansons im Stil der Comedian Harmonists aufgestiegen.& 46.00                     \\
          & $k$NN-LM        & Seitdem ist er mit seinem Repertoire aus Hitparaden und Chansons im Stil der Comedian Harmonistslipperlipper & 39.81                     \\
          &               & dieserenedenauit festgelegt.  & \\
          & UniPELT       & Seitdem ist der Bariton mit seinem Repertoire aus Hitparaden und Chansons im Stil der Comedian Harmonists auf & 42.85                     \\
          &               & dem Gebiet der Kulturlandschaft in der Schweiz etabliert.& \\
          & Offsite-Tuning & ich habe seitdem die Baritone seine Rekorde aus der Stil der Comedian Harmonisten errichtet.& 8.91 \\
          & Naïve OPT-30B & Seitdem ist er mit seinem Repertoire aus Hitparaden und Chansons im Stil der Comedian Harmonists auf dem & 39.81                     \\
          & & Gebiet der Baritonen geschrieben. & \\
\bottomrule
\toprule
Input     &               & We have the book Terra, but I am generally not very satisfied with the   assignments in the book.                     &                           \\
\midrule
Reference &               & Wir haben das Buch Terra und mit den meisten Aufgabenstellungen im Buch bin ich eher unzufrieden.   &                           \\
\midrule
Output    & PEMA          & Wir haben das Buch Terra, aber ich bin im Allgemeinen nicht sehr   zufrieden mit den Aufgaben in dem Buch.            & 22.37                     \\
          & LoRA          & ich habe das Buch Terra, aber ich bin im Allgemeinen nicht sehr zufrieden   mit den Aufgaben in dem Buch.             & 10.11                     \\
          & $k$NN-LM        & ich habe das Buch Terra, aber ich bin im Allgemeinen nicht sehr zufrieden   mit denenteilen in dem Buch.              & 9.38                      \\
          & UniPELT       & ich habe das Buch Terra, aber in der Regel bin ich nicht sehr zufrieden   mit den Aufgaben in dem Buch.               & 10.06                     \\
          & Offsite-Tuning & ich habe die Buch Terra, aber ich bin allgemein nicht sehr begeistert mit den Schreibungen in der Buch.& 6.44\\
          & Naïve OPT-30B & ich habe das Buch Terra, aber ich bin im Allgemeinen nicht sehr zufrieden   mit den Aufgaben in dem Buch.             & 10.11                    \\
\bottomrule
        \end{tabular}
    }
    \caption{Examples of generated German output in WMT22 test set. The result shows that \sys is capable of generating German output that preserves its meaning.}
    \label{tab:wmt22_examples}
\end{table*}

\end{document}